\documentclass{article}
\usepackage{spconf,amsmath,graphicx,hyperref}
\usepackage{amssymb, multirow} 

\title{ROBUST PRIOR-GUIDED SEGMENTATION FOR EDITABLE 3D GAUSSIAN SPLATTING}
\name{Raushan Joshi, Jean-Yves Guillemaut}
\address{University of Surrey, Guildford, Surrey, UK\\
         }
\begin{document}
%
\maketitle
\begin{abstract}
3D Gaussian Splatting (3D-GS) enables real-time 3D scene reconstruction but lacks robust segmentation for editing tasks such as object removal, extraction, and recoloring. Existing approaches that lift 2D segmentations to the 3D domain scene suffer from view inconsistencies and coarse masks. In this paper, we propose a novel framework that leverages the Segment Anything Model – High Quality (SAM-HQ) to generate accurate 2D masks, addressing the limitations of the standard SAM in boundary fidelity and fine-structure preservation. To achieve robust 3D segmentation of any target object in a given scene, we introduce a prior-guided label reassignment method that assigns labels to 3D Gaussians by enforcing multiview consistency with learned priors. Our approach achieves state-of-the-art segmentation accuracy and enables interactive, real-time object editing while maintaining high visual fidelity. Qualitative results demonstrate superior boundary preservation and practical utility in Virtual Reality (VR) and robotics, advancing 3D scene editing.
\end{abstract}
\section{Introduction}
\label{sec:intro}
Accurate representation and editing of three-dimensional (3D) scenes are central problems in computer vision and graphics, with applications~\cite{carmigniani2011augmented} such as virtual and augmented reality (VR/AR), autonomous navigation, robotics, and 3D content creation. Realistic 3D reconstruction enables photorealistic rendering from novel viewpoints and provides the basis for downstream tasks such as object detection, scene understanding, and interactive editing. Neural Radiance Fields (NeRFs)~\cite{mildenhall2021nerf} demonstrated high-fidelity novel view synthesis but are implicit representations and computationally expensive, requiring hours to days for training and rendering, which limits real-time usability. Recently, 3D Gaussian Splatting (3D-GS)~\cite{kerbl20233d} has emerged as a significant advancement, representing scenes with explicit 3D Gaussian primitives parameterized by position, scale, rotation, opacity, and color. Despite these advantages, standard 3D-GS lacks native semantic segmentation or object-level decomposition, constraining its utility for scene editing tasks. 3D scene segmentation remains under-explored compared to 2D segmentation due to limited 3D datasets~\cite{Yang2024SAMPart3DSA} and the high computational cost of processing point clouds. 

Recent methods~\cite{cen2025segment, ye2025gaussiangrouping} leverage 2D segmentation models like DINO~\cite{Caron_2021_ICCV} and SAM~\cite{Kirillov_2023_ICCV} to enhance 3D scene understanding through feature field distillation or 2D mask lifting. However, feature distillation often yields aliased masks with jagged edges due to low-resolution 2D features. Mask-lifting strategies~\cite{Zhu_2025_CVPR, ye2025gaussiangrouping} project 2D SAM masks into 3D scene and assign identity encodings to Gaussians, but these approaches can be computationally expensive and tend to suffer from mask quality, noise amplification, and poor adaptability to unseen viewpoints.
In this work, we address segmentation quality in 3D-GS by leveraging high-precision masks from SAM-HQ~\cite{ke2023segment} together with a noise-filtering module to suppress artifacts and incomplete boundaries. Moreover, we observe that relying exclusively on labels learned during 3D-GS optimization often leads to suboptimal local minima~\cite{hu2024sagd}. Recent methods~\cite{hu2024sagd, shen2025flashsplat} formulate label assignment as a linear programming problem that assigns optimal Gaussian labels by pixel-overlap with 2D masks, but the need of heuristic background bias in these formulations reduces robustness in scenes with occlusions or fine structures. 

To overcome these limitations, we introduce a novel prior–guided label reassignment strategy that enriches the linear programming framework with scene-specific label priors which are trained jointly with 3D-GS optimization. This eliminates heuristic bias dependency, improves multiview consistency, and yields more precise object-level editing in 3D-GS scenes. To summarize, our key contributions are: 1. We integrate SAM-HQ masks for high-quality 3D object segmentation in Gaussian Splatting, to the best of our knowledge, for the first time. 2. We propose a robust prior-guided label reassignment method that circumvents heuristic background biases by optimally assigning labels to each 3D Gaussians. 3. We validate our method's superior performance on our new high-quality multiview mask dataset.

\begin{figure*}[!t]
\centering
\includegraphics[width=0.99\textwidth]{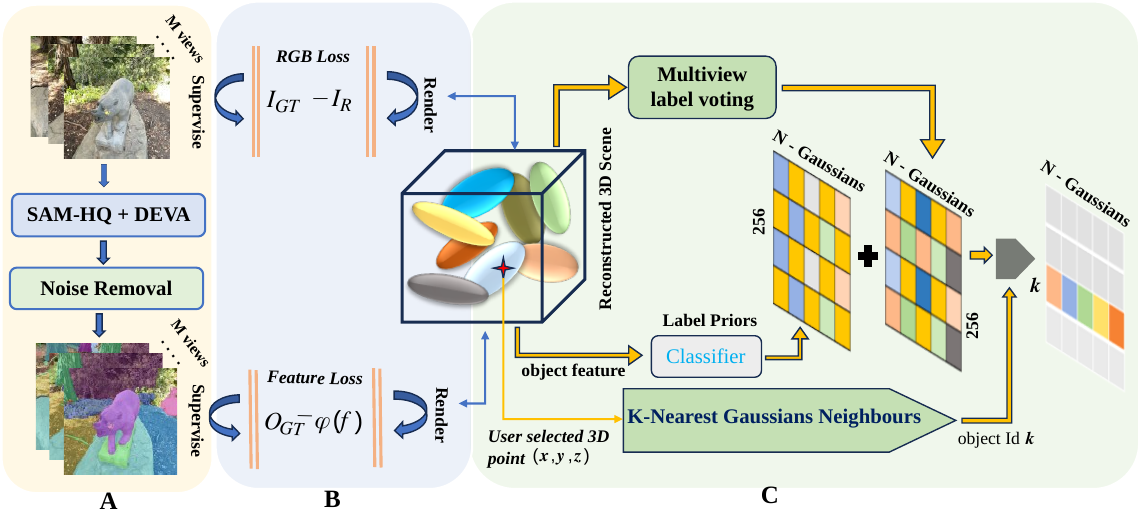}  
\caption{Overview of our robust 3D-GS segmentation framework. (\textbf{A}) Multiview mask generation with SAM-HQ for precise extraction and DEVA for consistent labelling via bidirectional temporal propagation. (\textbf{B}) Training: 16 dimensional per-Gaussian object features optimized jointly, projected via $\phi(f)$ to 256 class probabilities, supervised by true labels for label-aware rendering. (\textbf{C}) Inference for label reassignment: From user selected point $(x,y,z)$, query $K$-nearest Gaussian neighbours, project Gaussians to masks for majority voting with priors to form contribution matrix \(R^{256 \times N}\), and extract binary mask for object $k$.}

\label{fig:architecture}
\end{figure*}

\section{Methods}
Given a set of input images of a scene \(\{I_t \mid t = 1, 2, \ldots, M\}\), our goal is to obtain a 3D Gaussian Splatting (3D-GS) framework that supports accurate object-level editing. We first generate high-quality 2D masks for each image using SAM-HQ~\cite{ke2023segment} with DEVA~\cite{Cheng_2023_ICCV}, which provide reliable boundaries and object separation, as shown in part \textbf{A} of our proposed framework architecture in Fig.~\ref{fig:architecture}. These masks are used as supervision during the joint training of 3D-GS, ensuring that the learned Gaussian primitives are aligned with object features (part \textbf{B} in Fig.~\ref{fig:architecture}). Next, we propose a prior-guided label reassignment strategy that combines majority voting across multiview Gaussians with learned object priors (part \textbf{C} in Fig.~\ref{fig:architecture}). This ensures consistent and optimal label assignment for each 3D Gaussian.

\subsection{High Quality Object Mask Generation}
\label{sec:hq_mask_generation}
\begin{figure}[t!]

\begin{minipage}[b]{1.0\linewidth}
\centering
\centerline{\includegraphics[width=7.5cm, height=3.5cm]{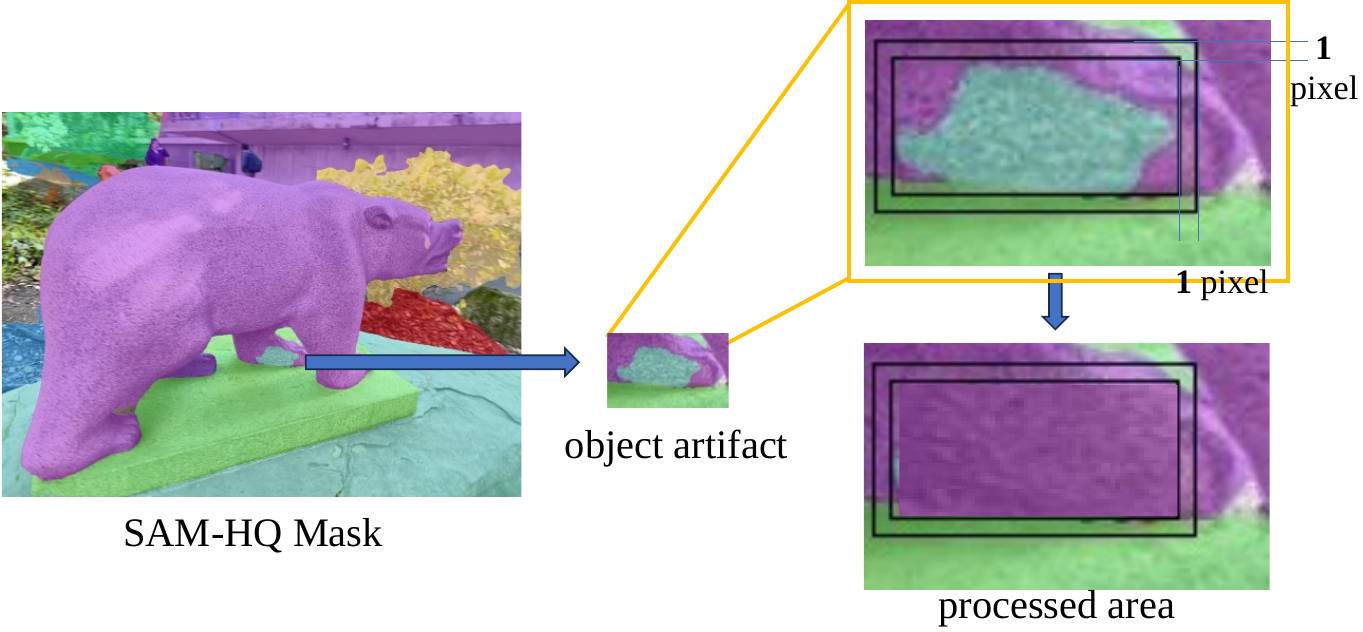}}
\end{minipage}
\caption{Overview of preprocessing step to remove small artifacts using connected component labeling.}
\label{fig:preprocessing}
\end{figure}

To initiate our methodology, we employ the SAM-HQ~\cite{ke2023segment} model to generate high-quality segmentation masks for multiview images of the scene, leveraging its enhanced precision and ability to produce sharper masks for complex structures compared to standard methods~\cite{Kirillov_2023_ICCV}. However, applying SAM-HQ independently to each view risks producing inconsistent object labels. To address this, the Decoupled Video Segmentation method (DEVA)~\cite{Cheng_2023_ICCV} combines SAM-HQ’s image-level segmentation with a class-agnostic temporal propagation model, treating multiview images as video frames to ensure consistent labelling. DEVA’s bidirectional propagation, also used in previous works~\cite{ye2025gaussiangrouping}, employs a memory bank to track objects via feature similarity and spatial coherence. To further refine the masks, a post-processing pipeline applies morphological closing~\cite{gonzalez2008digital} with a 3×3 square kernel to smooth jagged edges and reduce noise-induced inconsistencies. Additionally, small objects below the pixel area threshold (\textit{set to 500 by default}) are reassigned to dominant surrounding labels using connected component analysis~\cite{gonzalez2008digital}. Fig.~\ref{fig:preprocessing} shows how the method enhances mask quality, providing robust supervision for joint training of object features in 3D-GS scene.

\subsection{Object Feature Based 3D Gaussian Splatting}
3D Gaussian Splatting (3D-GS)~\cite{kerbl20233d} is an efficient method for real-time 3D scene representation and rendering, offering high-quality novel view synthesis. It represents a 3D scene as a set of explicit 3D Gaussians, each parameterized by a center position \( \mu \in \mathbb{R}^3 \), a covariance matrix \( \Sigma \) defined by scaling \( s \in \mathbb{R}^3 \) and rotation quaternion \( q \in \mathbb{R}^4 \), opacity \( \alpha \in \mathbb{R} \), and color \( c \) modelled via spherical harmonics. The 3D Gaussian function is defined as:
\begin{equation}
G(x) = \frac{1}{(2\pi)^{3/2}|\Sigma|^{1/2}} e^{-\frac{1}{2}(x-\mu)^T \Sigma^{-1} (x-\mu)}.
\end{equation}

These 3D Gaussians are projected onto the 2D image plane as 2D Gaussians, rendered using a differentiable tile-based rasterization pipeline~\cite{kerbl20233d}. For each pixel \(u\), the rendered color \( C(u)\) is computed via \(\alpha\)-blending:
\begin{equation}
C(u) = \sum_{i \in N} c_i \alpha_i \prod_{j=1}^{i-1} (1 - \alpha_j) = \sum_{i \in N} c_i \alpha_i T_i
\label{eq:rasterization_equation}
\end{equation}
where \( \alpha_i = o_i G'_i(u) \), \(o_i\) is the opacity, \(T_i\) is the transmittance, and \( G'_i \) represents the projected 2D Gaussian splat. 

In this paper, we augment each Gaussian with an additional 16-dimensional view-invariant object feature vector \( f_{o,i} \in \mathbb{R}^{16} \), encoding object IDs for up to 256 unique objects. The 16-dimensional feature size is selected to balance computational efficiency with sufficient capacity to represent a large number of objects. View invariance is enforced by setting the spherical harmonics (SH) degree of $f_{o,i}$ to zero, ensuring that the object ID remains consistent regardless of the viewing angle.

Further, we leverage the 3D-GS differentiable rasterization pipeline~\cite{kerbl20233d} to retain efficient rendering for object features:
\begin{equation}
 F_o(u) = \sum_{i \in N} f_{o,i} \alpha_i \prod_{j=1}^{i-1} (1 - \alpha_j)
\end{equation}
where \( F_o(u) \in \mathbb{R}^{16} \) is object feature rendered at pixel \(u\). The rendered object feature \( F_o \) is then passed through a trainable linear layer \( \phi_L: \mathbb{R}^{16} \to \mathbb{R}^{256} \), producing logits \( Z = \phi_L(F_o) \) and, later a softmax function converts logits \( Z \) into probabilities over 256 object classes:
\begin{equation}
P(k) = \frac{\exp(Z_k)}{\sum_{j=0}^{255} \exp(Z_j)}.
\label{object_probability}
\end{equation}

The object features and linear layer parameters are optimized using a cross-entropy loss:
\begin{equation}
L_{\text{obj}} = -\frac{1}{N_p} \sum_{p=1}^{N_p} \sum_{k=0}^{255} \mathbb{I}\{y_p = k\} \log(P_p(k))
\label{cross_entropy_loss}
\end{equation}
where \( N_p \) is the number of pixels, $\mathbb{I}$ is an indicator function, \( y_p \in [0, 255] \) is the ground-truth object ID from SAM-HQ masks~\cite{ke2023segment}, and \( P_p(k) \) is the predicted probability for class \( k \) at pixel \( p \). This loss, effective for multiclass classification~\cite{gonzalez2008digital}, encourages accurate object mask assignments. The overall training objective combines the segmentation loss with the standard 3D-GS photometric loss \( L_{\text{rgb}} \)~\cite{kerbl20233d} as follows:
\begin{equation}
L_{\text{total}} = L_{\text{rgb}} + \gamma_{\text{obj}} L_{\text{obj}}
\label{total_loss}
\end{equation}
where \( \gamma_{\text{obj}} \) balances RGB reconstruction and object segmentation accuracy. This formulation ensures efficient 3D-GS segmentation supervised by high-quality 2D masks.

\subsection{Prior-Guided Label Reassignment}
Having trained object features in the 3D-GS scene, we need to enhance robustness against noisy 2D masks. We formulate label reassignment as a linear programming problem~\cite{shen2025flashsplat}, where Gaussians that significantly contribute to particular object ID \(l\) in all given masks are designated with label \(l\) using simple majority voting. Here, we improve the contribution score for each Gaussian to include a regularization term as follows:
\begin{equation}
A^{\text{new}}_{l,i} = A_{l,i} + \gamma_p \cdot p_{i,l_i} \cdot \mathbb{I}\{l_i = l\}
\label{eq:new_label_score}
\end{equation}
\begin{equation}
A_{l,i} = \sum_{v,j,k} \alpha_i(j,k) \cdot T_i(j,k) \cdot \mathbb{I}(M_v(j,k) = l)
\label{eq:flashsplat_label}
\end{equation}
where \( \gamma_p \) controls regularization strength, \( p_{i,l_i} \) is the prior score (softmax probabilities from Eq.~\ref{object_probability}) for Gaussian’s label \( l_i \), \( \alpha_i\) and \(T_i\) are constant during rendering (from Eq.~\ref{eq:rasterization_equation}), and \( \mathbb{I}\{M_v(j, k) = l\} \) indicates if the ground-truth mask \( M_v \) assigns the label \( l \) to pixel \( (j, k) \). The new label to \(i\)-th Gaussian is assigned as:
\begin{equation}
P^{\text{new}}_i = \arg\max_l A^{\text{new}}_{l,i}.
\label{eq:final_label_argmax}
\end{equation}

This approach integrates learned priors with a majority voting method, thus eliminating heuristic-based bias. The resulting segmentation is represented as a matrix \( S \in \mathbb{R}^{256 \times N} \), where \( S_{l,n} \in \{0, 1\} \) indicates if Gaussian \( n \) belongs to class \( l \). Our method improves multiview consistency and robustness, outperforming traditional label voting techniques.

For inference, given a user-selected 3D point \( p \in \mathbb{R}^3 \) in the 3D scene, we identify the \( K \)-nearest Gaussians, each with a predicted object label \( \ell_i \in [0, 255] \). The mask ID \(k\) for the object is predicted via majority voting as follows:
\begin{equation}
k = \text{Mode}(\{\ell_1, \ell_2, \dots, \ell_K\}).
\end{equation}
Later, using prior-guided label reassignment defined in Eq.~\ref{eq:new_label_score}, the Gaussian binary mask for the selected object is extracted by identifying Gaussians that contribute maximally to label \( k \), while leveraging local spatial consistency for robustness.

\section{Experiments}

\subsection{Experimental Settings}
\textbf{Datasets and Metrics:} We evaluate the performance on datasets containing synthetic and real-world scenes such as LeRF~\cite{Kerr_2023_ICCV}, Mip-NeRF~\cite{Barron_2022_CVPR} and LLFF~\cite{mildenhall2021nerf, ye2025gaussiangrouping}. Further, by leveraging SAM-HQ’s~\cite{ke2023segment} high-quality output in our preprocessing pipeline, we establish a new high-quality object mask dataset. The proposed high-quality object mask dataset consists of five representative scenes: \textit{bear}, \textit{ramen}, \textit{horns}, \textit{teatime}, and \textit{fortress}. These scenes were carefully selected from the \textit{Mip-NeRF 360} and \textit{LLFF} benchmarks due to their complexity, enabling a thorough evaluation of our pipeline under realistic conditions on a mid-range GPU. All masks were generated using SAM-HQ, followed by our noise removal preprocessing pipeline, and were subsequently manually verified for boundary accuracy and label consistency, as described in Section~\ref{sec:hq_mask_generation}. The complete dataset is publicly available at:
\url{https://huggingface.co/datasets/joshir/3D-Scene-Segmentation-HQ}. We evaluated rendering quality using PSNR~\cite{wang2020image}, SSIM~\cite{chen2011fast}, and LPIPS~\cite{snell2017learning} metrics. For segmentation and scene editing tasks such as object removal, we use the mean Intersection over Union (mIoU) and mean Accuracy (mAcc) metrics for evaluation.

\textbf{Implementation Details:} We first train object features with the official 3D-GS codebase~\cite{kerbl20233d}. All parameters are jointly optimized using the Adam~\cite{Kingma2014AdamAM} optimizer with learning rates of 0.005 for object features and 0.0005 for the linear layer, over 30K iterations across all scenes. For label reassignment, the contribution matrix $\mathbf{S} \in \mathbb{R}^{256 \times L}$ is computed using tile-based rasterization. The default prior weight $\gamma_{\text{p}}$ is set to 0.2. The entire 3D-GS training and label reassignment method is executed on NVIDIA RTX~A4000 GPU.
\begin{table}[t!]
\renewcommand{\arraystretch}{1.0} 
\centering
\caption{Scene reconstruction fidelity comparison between our proposed approach based on SAM-HQ~\cite{ke2023segment} and standard SAM~\cite{Kirillov_2023_ICCV} masks.}
\vspace{1mm}
{\fontsize{9pt}{10pt}\selectfont
\begin{tabular}{l|cc|cc|cc}
\hline
\multirow{2}{*}{Scene} & \multicolumn{2}{c|}{\textbf{PSNR~$\uparrow$}} & \multicolumn{2}{c|}{\textbf{SSIM~$\uparrow$}} & \multicolumn{2}{c}{\textbf{LPIPS~$\downarrow$}} \\
&  SAM  &  Ours &  SAM  &  Ours & SAM  &  Ours  \\
\hline
bear & 28.53 & \textbf{28.83} & 0.907 & \textbf{0.908} & 0.136 & \textbf{0.134} \\
ramen & 27.71  & \textbf{28.49}  & 0.903 & \textbf{0.907} & 0.167 & \textbf{0.161} \\
horns &  23.70  & \textbf{24.11} & 0.851 & \textbf{0.863} & 0.216 & \textbf{0.181} \\
teatime & 30.09 & \textbf{30.42}  & 0.913 & \textbf{0.918} & 0.141 & \textbf{0.130} \\
fortress & 33.28 & \textbf{33.31} & 0.931 & \textbf{0.933} & 0.111 & 0.111 \\
\hline
\textbf{Overall} & 28.66 & \textbf{29.04} & 0.901 & \textbf{0.906} & 0.154 & \textbf{0.143}\\
\hline
\end{tabular}
\label{tab:reconstruction_metrics_results}
}
\end{table}
\begin{table}[t!]
\renewcommand{\arraystretch}{1.0} 
\centering
\caption{Quantitative segmentation results using point prompt on standard NVOS~\cite{Ren_2022_CVPR} dataset with 8 front-facing scenes.}
\vspace{1mm}
{\fontsize{9pt}{10pt}\selectfont
\begin{tabular}{l|c|c}
\hline
\multirow{1}{*}{\textbf{Method}} & \multicolumn{1}{c|}{\textbf{mIoU (\%)~$\uparrow$}} & \multicolumn{1}{c}{\textbf{mAcc (\%)~$\uparrow$}} \\

\hline
NVOS & 39.4 & 73.6 \\
ISRF & 70.1 & 92.0 \\
FlashSplat & \textbf{91.8} & \textbf{98.6} \\
\textbf{Ours} & 91.0 & 98.4 \\
\hline
\end{tabular}
\label{tab:method_result_comparison}
}
\end{table}
\begin{table}[!t]
\renewcommand{\arraystretch}{1.0} 
\centering
\caption{Segmentation comparison results between our proposed approach and Flashsplat~\cite{shen2025flashsplat} on our new high-quality object mask dataset.}
\vspace{1mm}
{\fontsize{9pt}{10pt}\selectfont
\begin{tabular}{l|cc|cc}
\hline
\multirow{2}{*}{\textbf{Scene}} & \multicolumn{2}{c|}{\textbf{mIoU (\%)~$\uparrow$}} & \multicolumn{2}{c}{\textbf{mAcc (\%)~$\uparrow$}} \\
&  Flashsplat  &  Ours &  Flashsplat  &  Ours  \\
\hline
bear & 88.73 & \textbf{89.58} & 97.07 & \textbf{97.36}\\
ramen & 68.49 & \textbf{69.69} & 99.59 & \textbf{99.63} \\
horns & \textbf{95.61} & 91.22 & \textbf{99.26} & 98.44\\
teatime & 77.61 & \textbf{78.3} & 98.24 & \textbf{98.81} \\
fortress & 90.50 & \textbf{91.03} & 98.45 & \textbf{98.62}\\
\hline
\textbf{Overall} & 80.25 & \textbf{80.52} & 98.55 & \textbf{98.57}\\
\hline
\end{tabular}
\label{tab:scene_wise_results_comparison}
}
\end{table}
\begin{figure}[t!]
\renewcommand{\arraystretch}{0.9}
\small
\centering
\hspace{0.9cm} \textbf{RGB}  \hspace{1.6cm}\textbf{SAM-HQ}  \hspace{1.2cm}\textbf{Ours} \\

\begin{minipage}{0.06\textwidth}
    \textbf{bear}
\end{minipage}
\begin{minipage}{0.13\textwidth}
    \includegraphics[width=\linewidth]{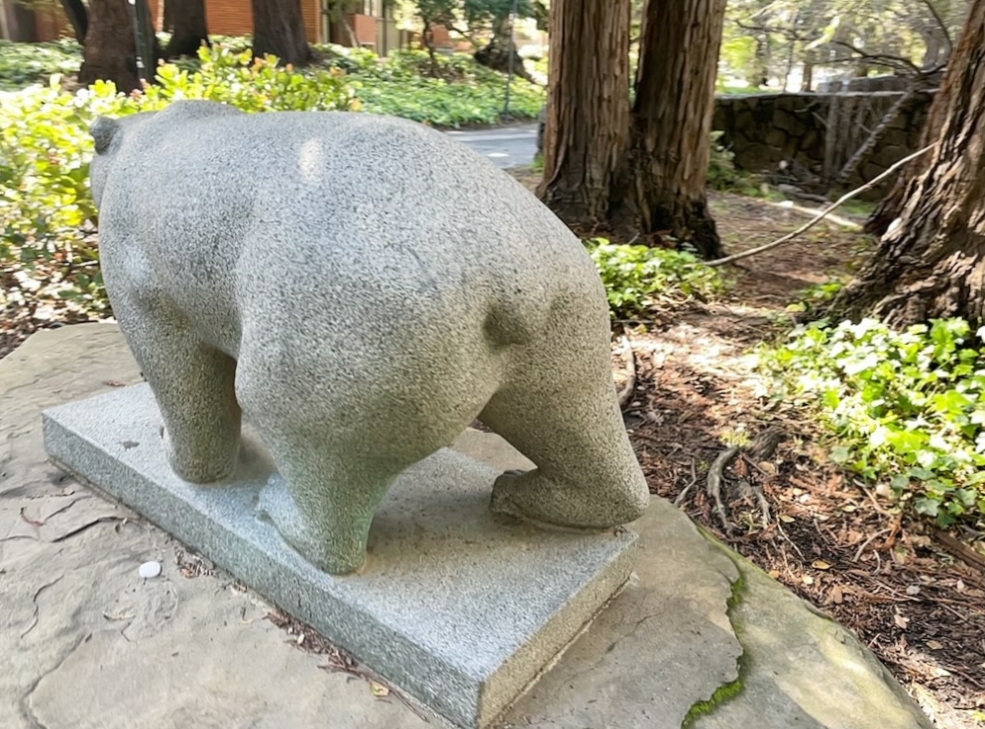}
\end{minipage} 
\begin{minipage}{0.14\textwidth}
    \includegraphics[width=\linewidth]{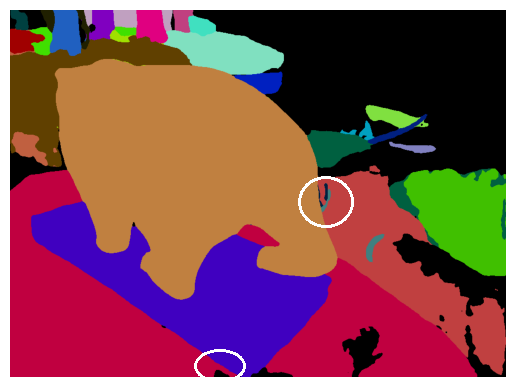}
\end{minipage} 
\begin{minipage}{0.14\textwidth}
    \includegraphics[width=\linewidth]{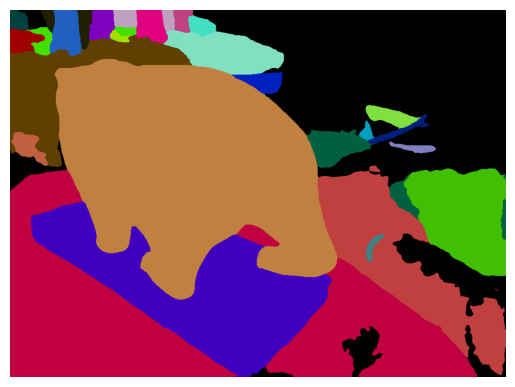}
\end{minipage} \\
\begin{minipage}{0.06\textwidth}
    \textbf{ramen}
\end{minipage}
\begin{minipage}{0.13\textwidth}
    \includegraphics[width=\linewidth]{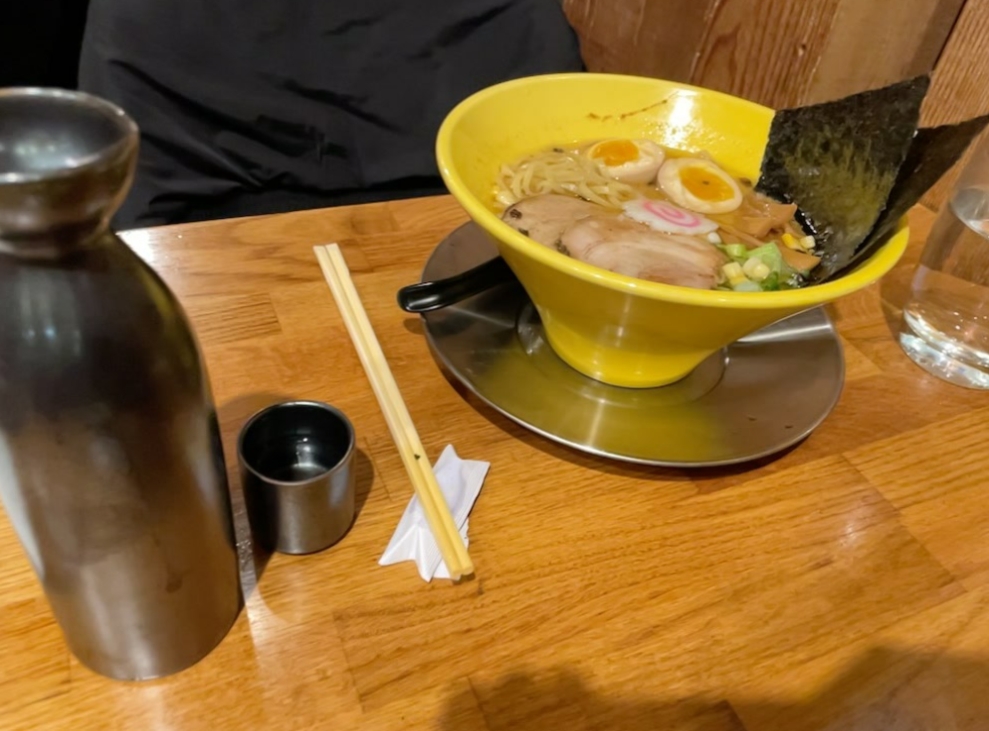}
\end{minipage} 
\begin{minipage}{0.14\textwidth}
    \includegraphics[width=\linewidth]{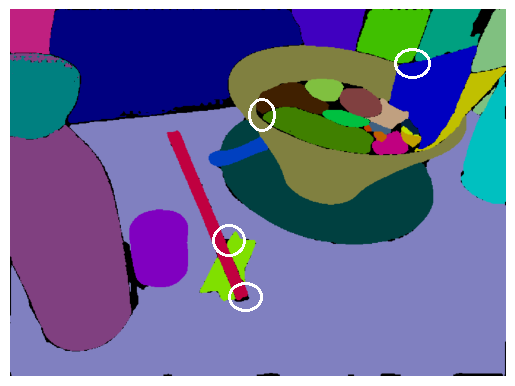}
\end{minipage} 
\begin{minipage}{0.14\textwidth}
    \includegraphics[width=\linewidth]{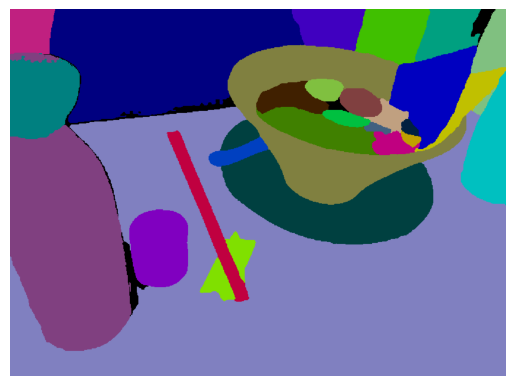}
\end{minipage} \\

\begin{minipage}{0.06\textwidth}
    \textbf{horns}
\end{minipage}
\begin{minipage}{0.13\textwidth}
    \includegraphics[width=\linewidth]{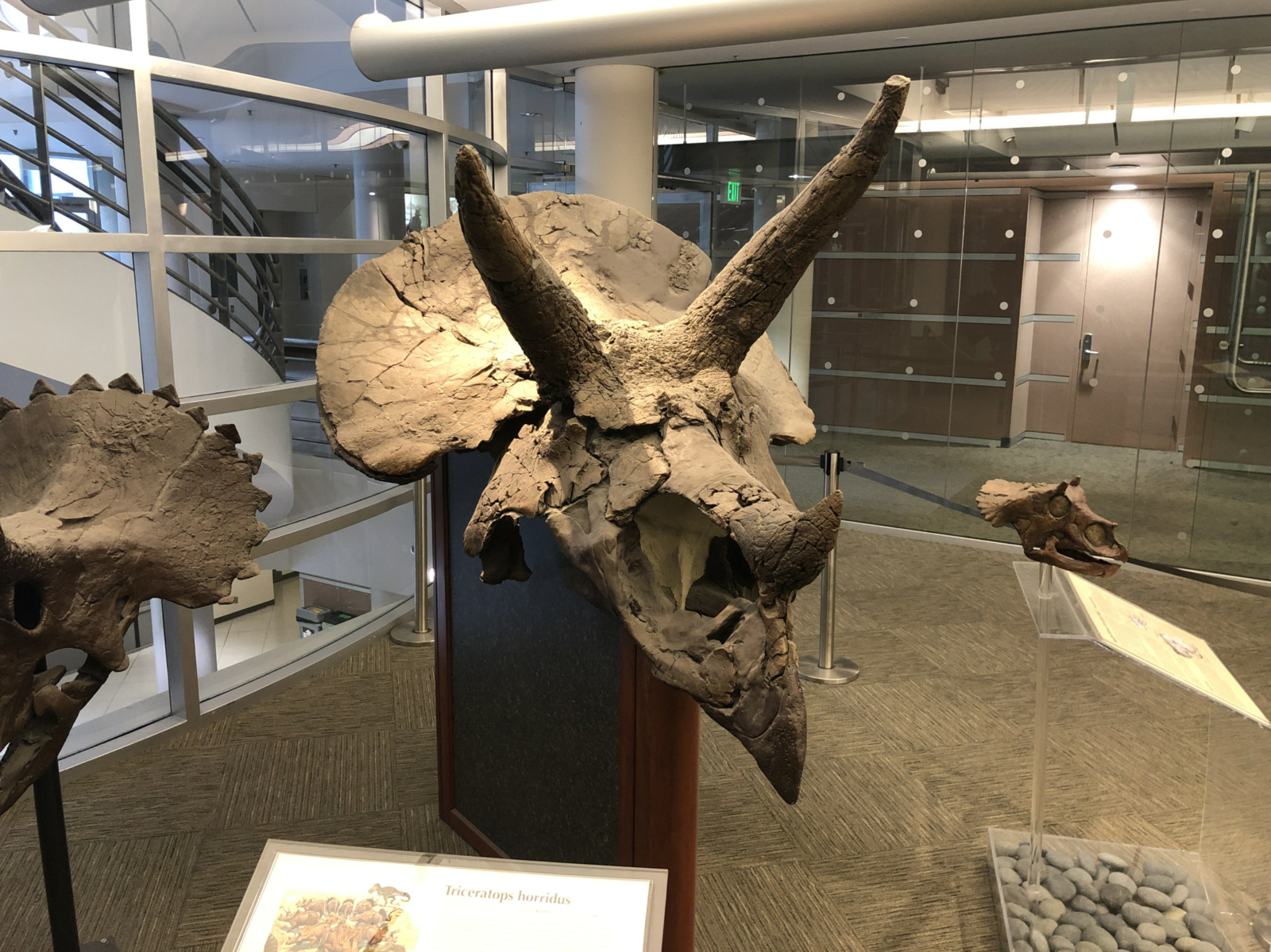}
\end{minipage} 
\begin{minipage}{0.14\textwidth}
    \includegraphics[width=\linewidth]{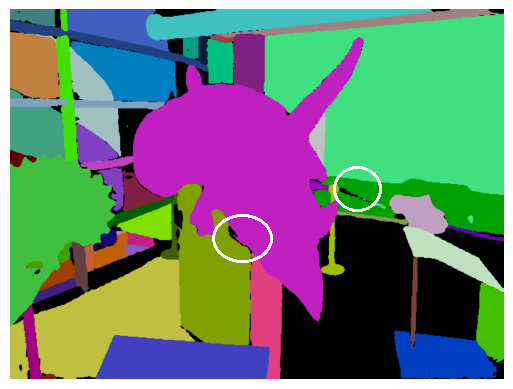}
\end{minipage}
\begin{minipage}{0.14\textwidth}
    \includegraphics[width=\linewidth]{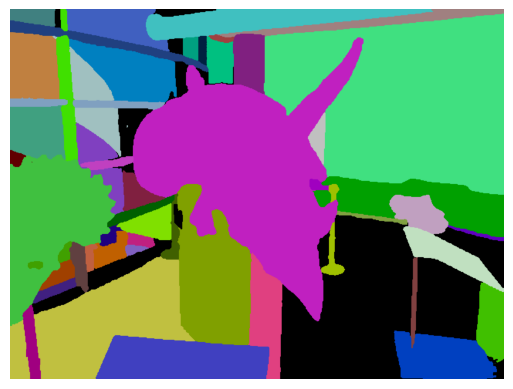}
\end{minipage}  \\
\begin{minipage}{0.06\textwidth}
    \textbf{teatime}
\end{minipage}
\begin{minipage}{0.13\textwidth}
    \includegraphics[width=\linewidth]{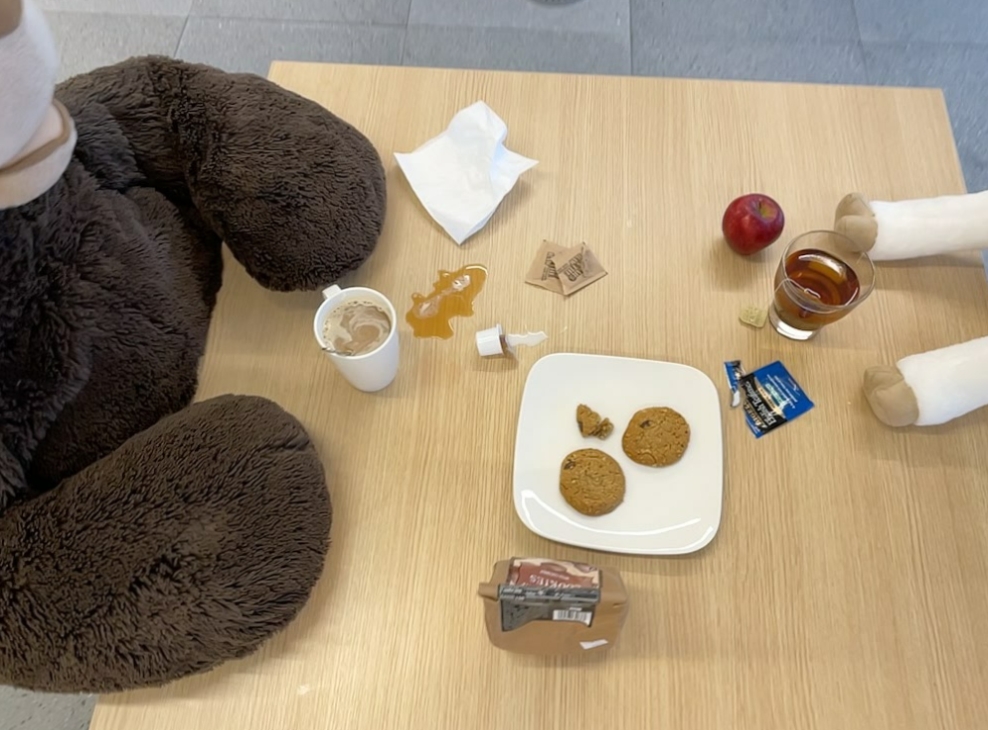}
\end{minipage} 
\begin{minipage}{0.14\textwidth}
    \includegraphics[width=\linewidth]{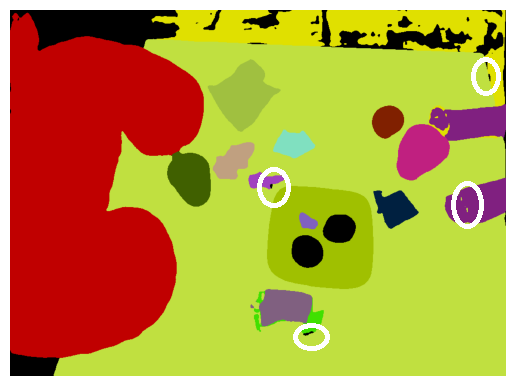}
\end{minipage} 
\begin{minipage}{0.14\textwidth}
    \includegraphics[width=\linewidth]{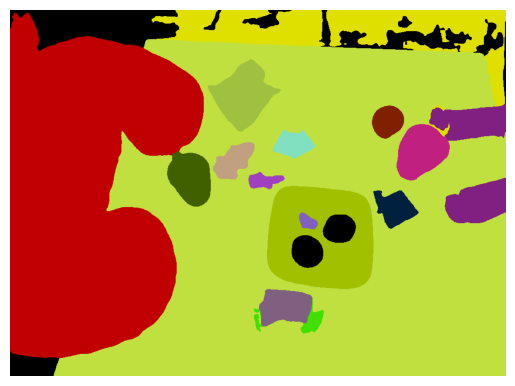}
\end{minipage} \\

\caption{Qualitative results of our Noise Removal module. \textbf{White circles} in the middle column show the artifacts.}
\label{fig:object_mask_extraction}
\end{figure}
\begin{figure}[t!]
\renewcommand{\arraystretch}{0.9}
\small
\centering
\hspace{0.8cm} \textbf{RGB} \hspace{1.1cm} \textbf{Recolor} \hspace{0.8cm} \textbf{Removal} \hspace{0.7cm} \textbf{Extraction} \hspace{0.1cm}\\[0.1cm]

\begin{minipage}{0.02\textwidth}
  \centering
  \rotatebox{90}{\textbf{bear}}
\end{minipage}
\begin{minipage}{0.11\textwidth}
    \includegraphics[width=\linewidth]{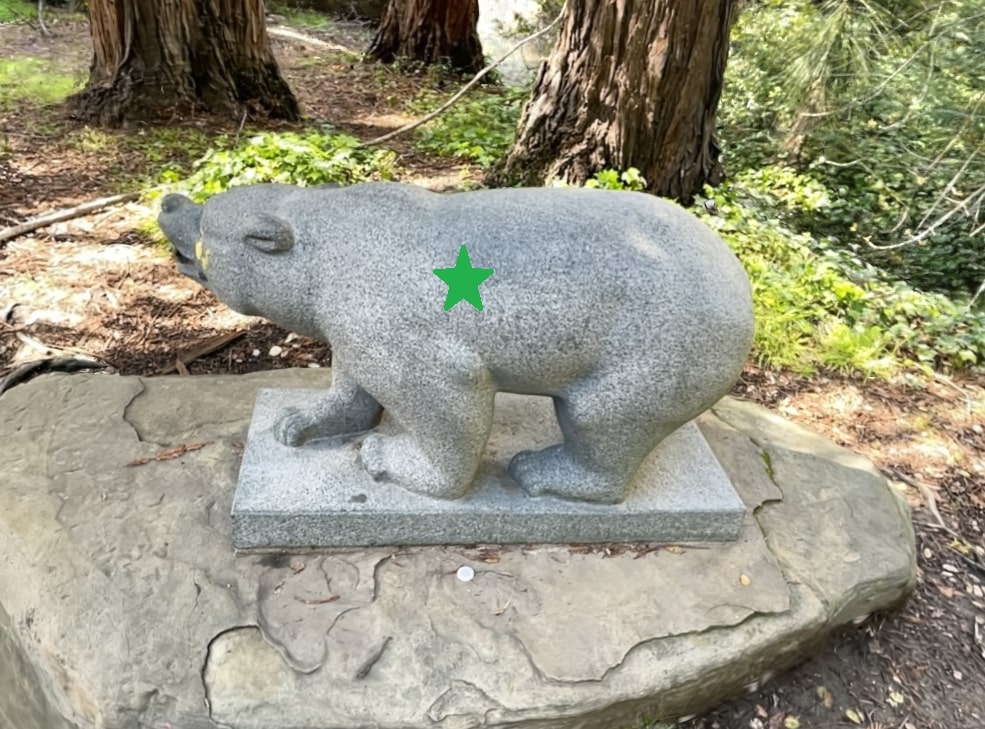}
\end{minipage}
\begin{minipage}{0.11\textwidth}
    \includegraphics[width=\linewidth]{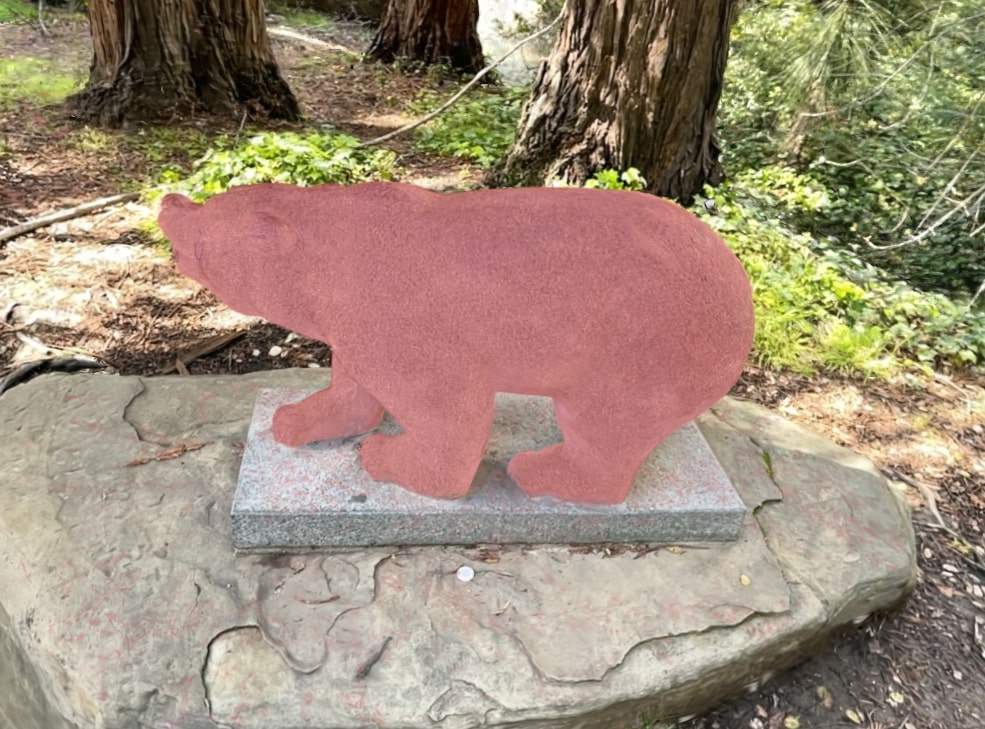}
\end{minipage}
\begin{minipage}{0.11\textwidth}
    \includegraphics[width=\linewidth]{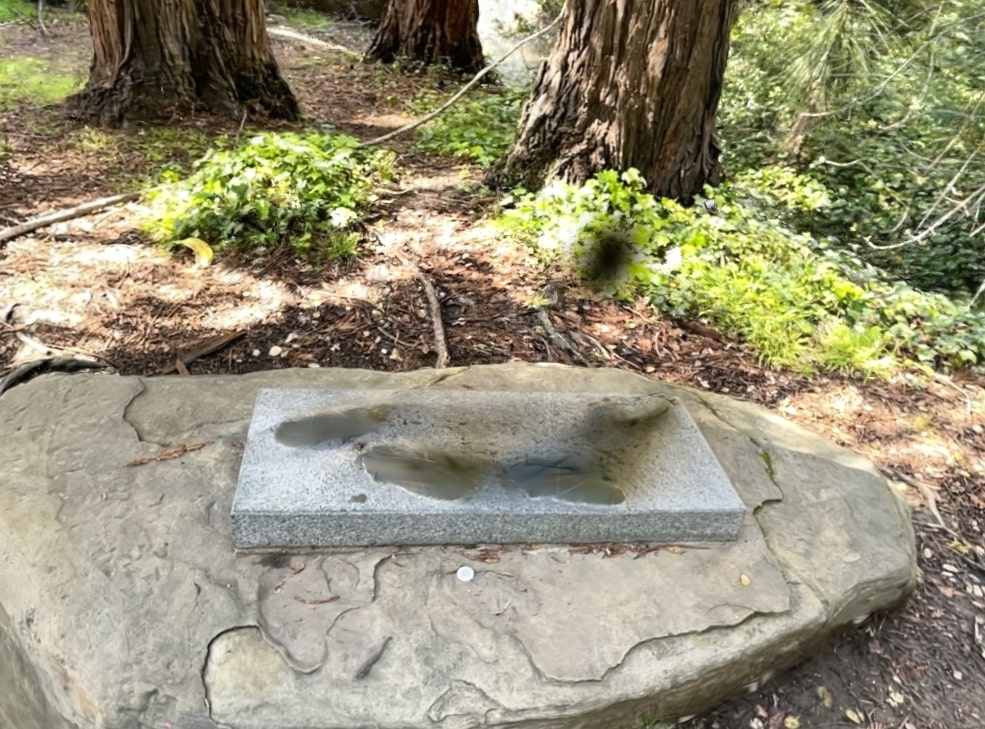}
\end{minipage} 
\begin{minipage}{0.11\textwidth}
    \includegraphics[width=\linewidth]{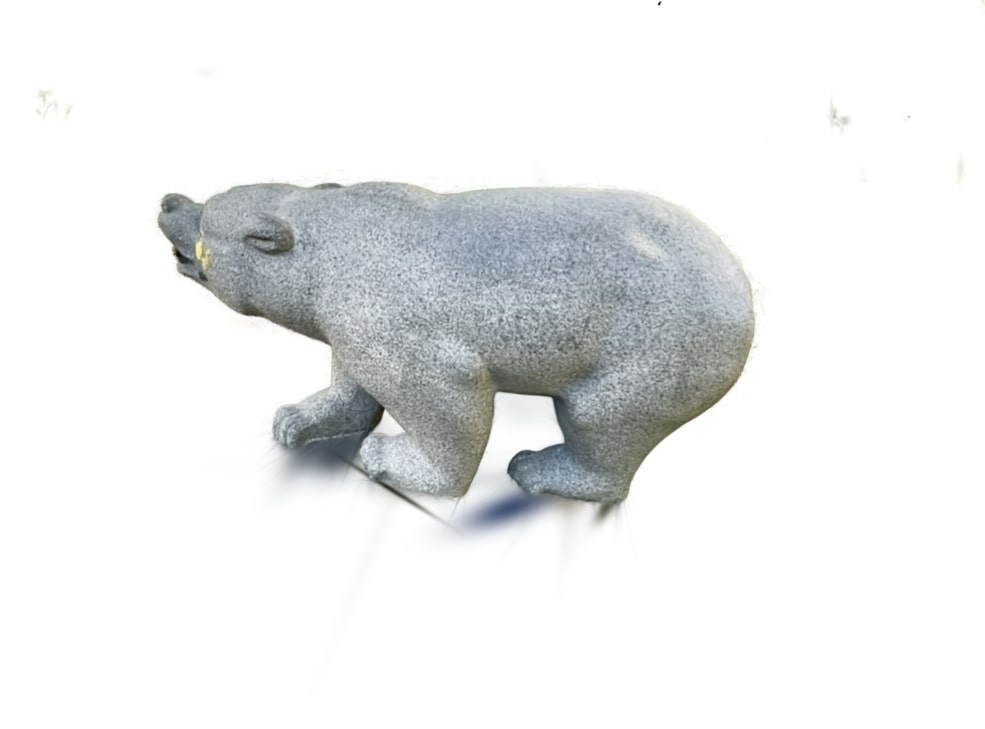}
\end{minipage} \\

\begin{minipage}{0.02\textwidth}
  \centering
  \rotatebox{90}{\textbf{ramen}}
\end{minipage}
\begin{minipage}{0.11\textwidth}
    \includegraphics[width=\linewidth]{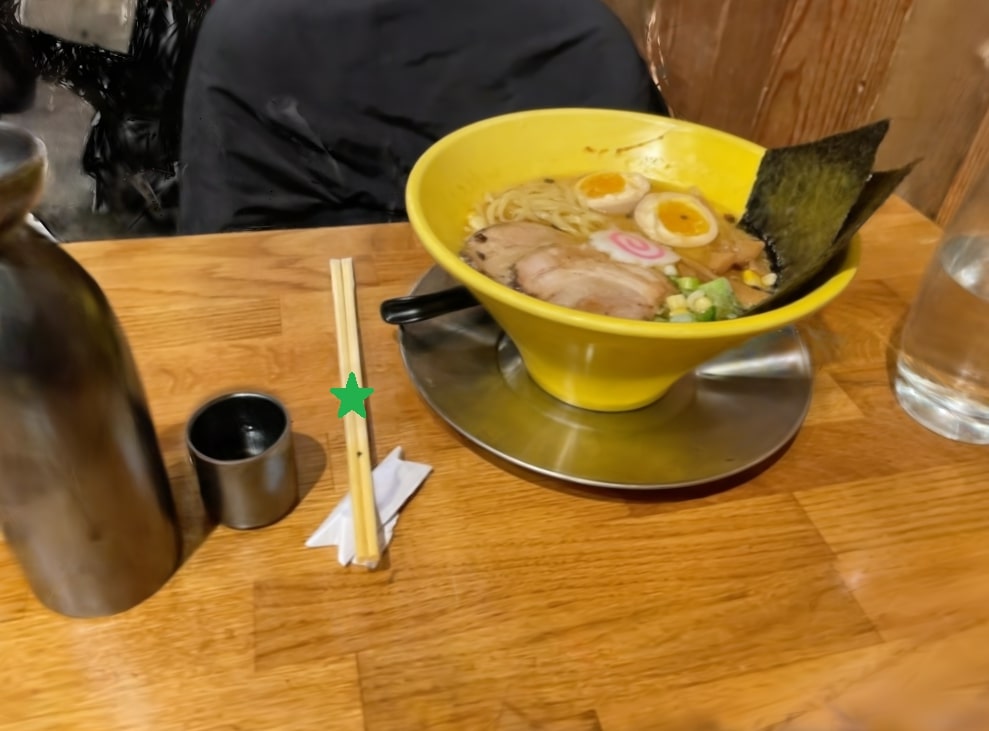}
\end{minipage}
\begin{minipage}{0.11\textwidth}
    \includegraphics[width=\linewidth]{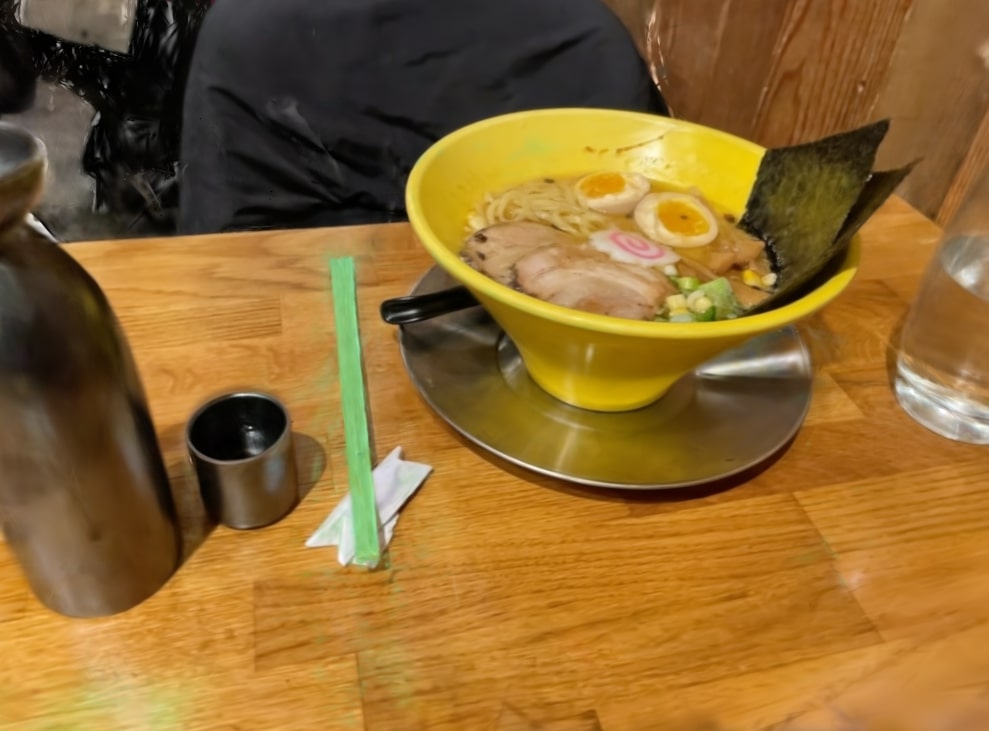}
\end{minipage}
\begin{minipage}{0.11\textwidth}
    \includegraphics[width=\linewidth]{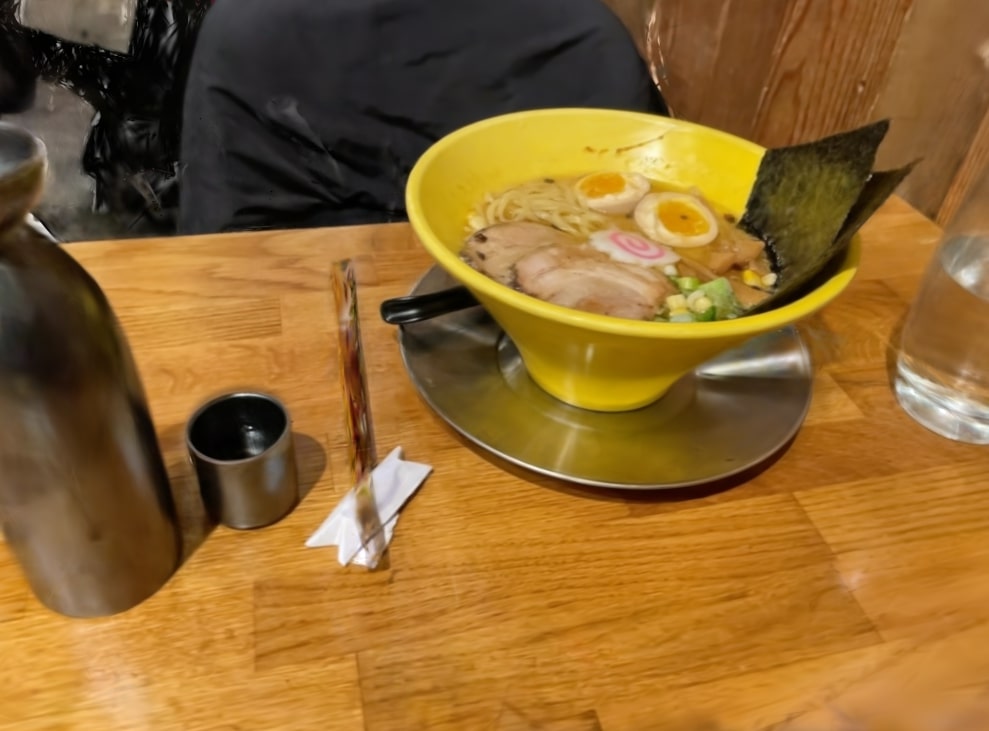}
\end{minipage}
\begin{minipage}{0.11\textwidth}
    \includegraphics[width=\linewidth]{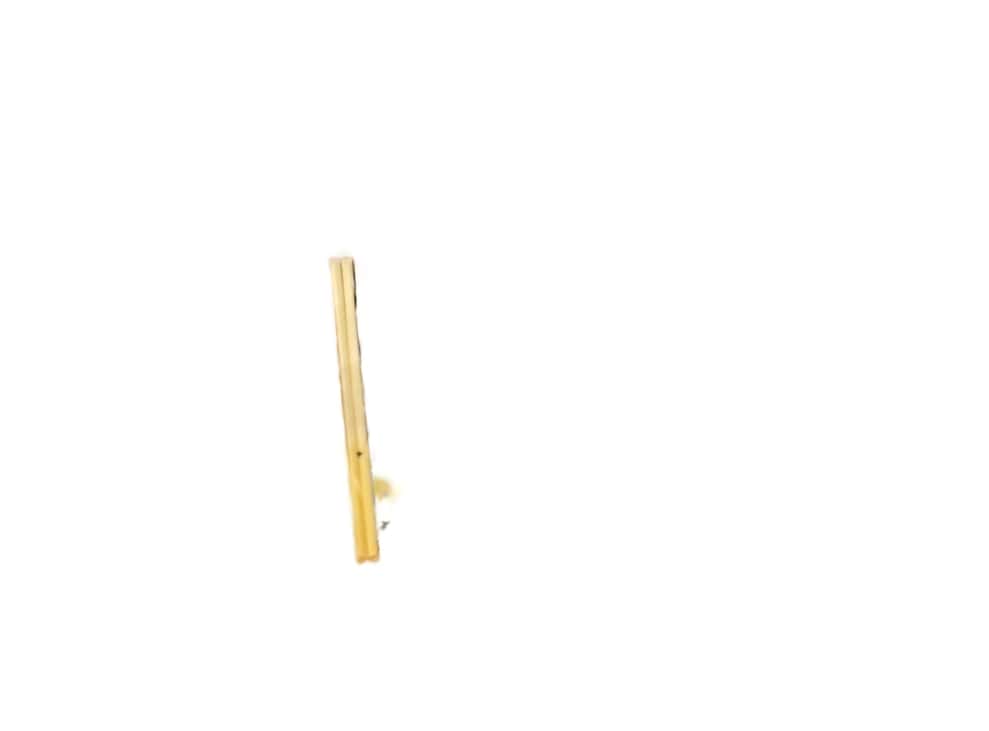}
\end{minipage}  \\

\begin{minipage}{0.02\textwidth}
  \centering
  \rotatebox{90}{\textbf{horns}}
\end{minipage}
\begin{minipage}{0.11\textwidth}
    \includegraphics[width=\linewidth]{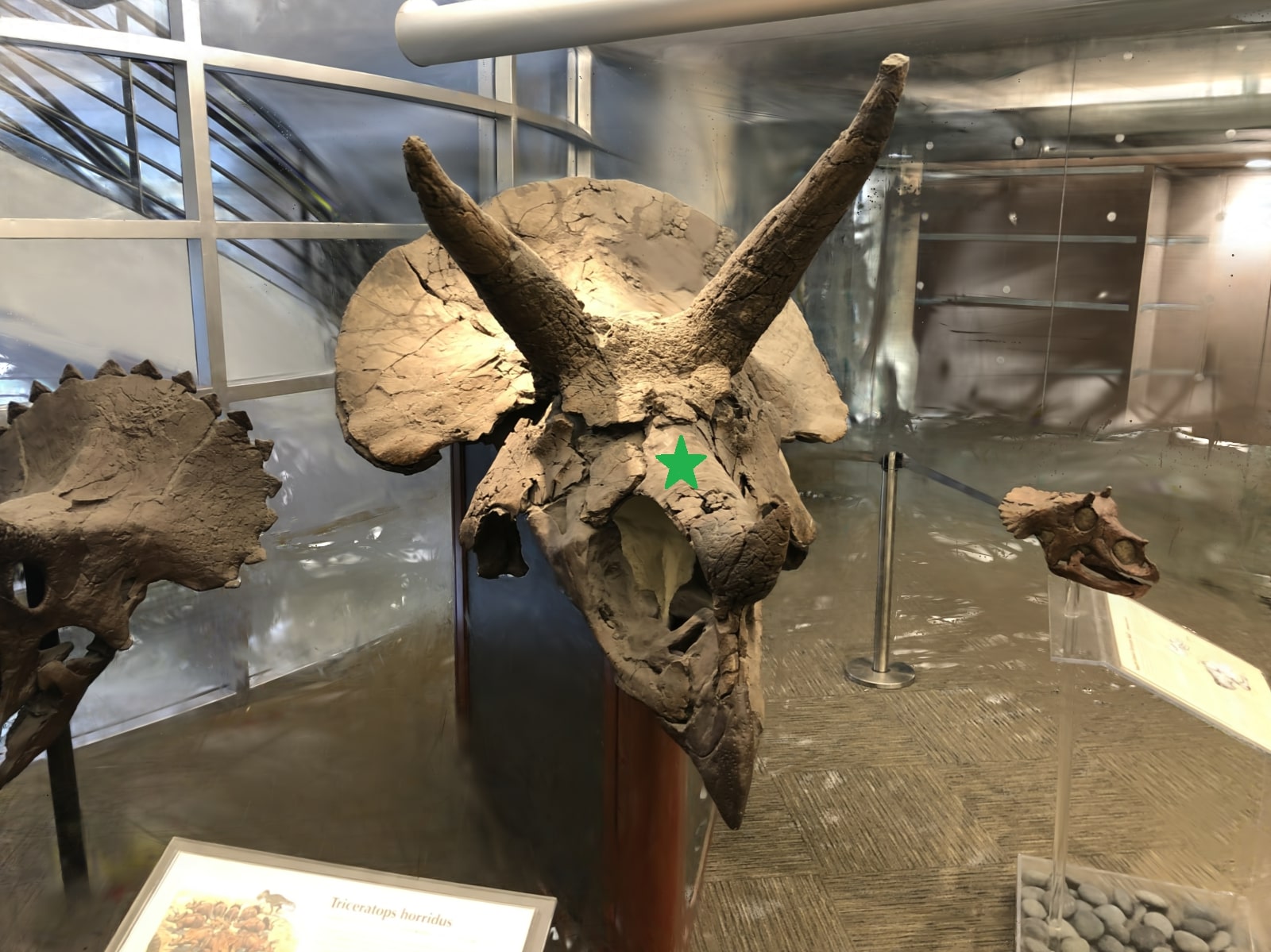}
\end{minipage}
\begin{minipage}{0.11\textwidth}
    \includegraphics[width=\linewidth]{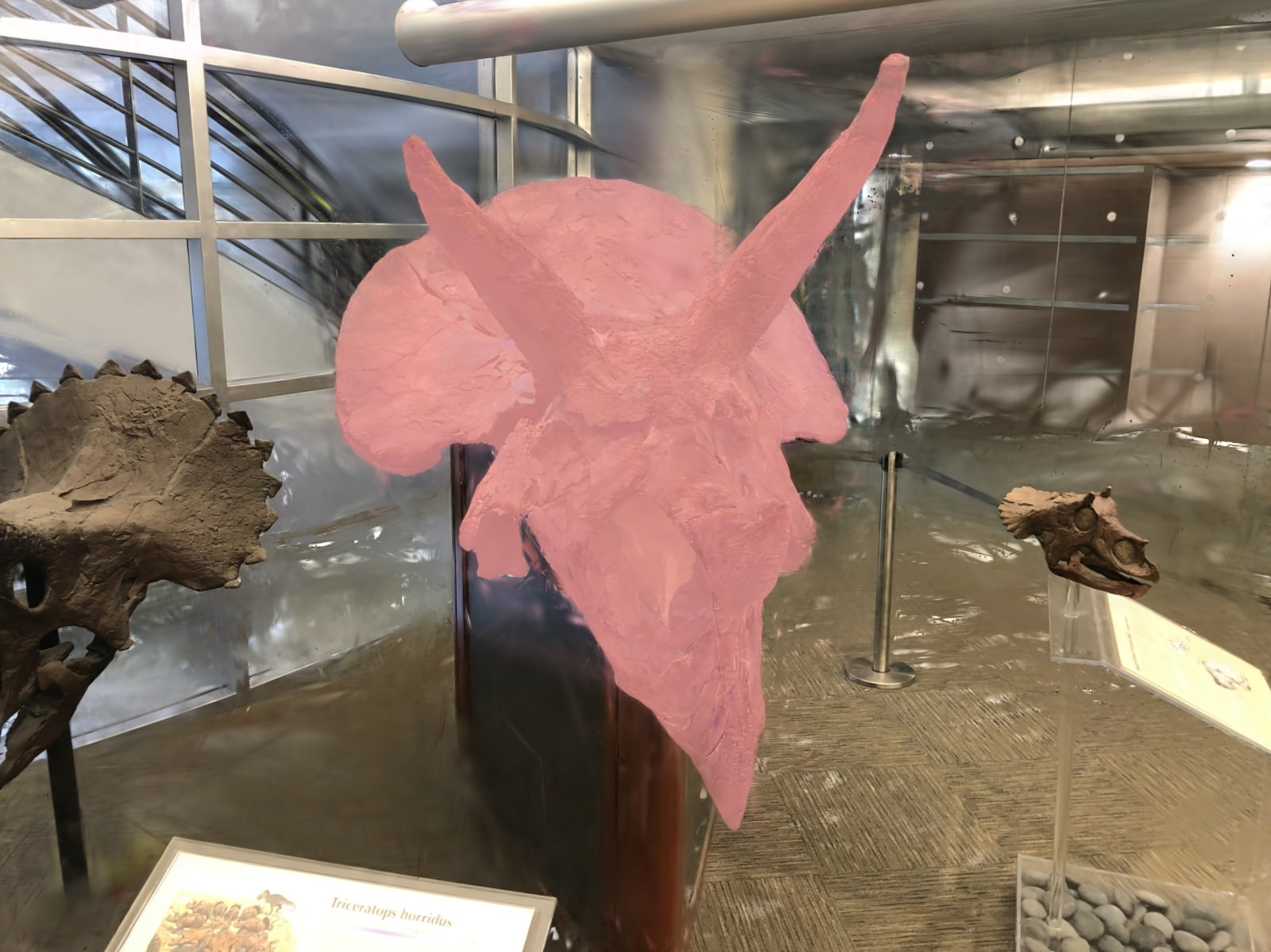}
\end{minipage}
\begin{minipage}{0.11\textwidth}
    \includegraphics[width=\linewidth]{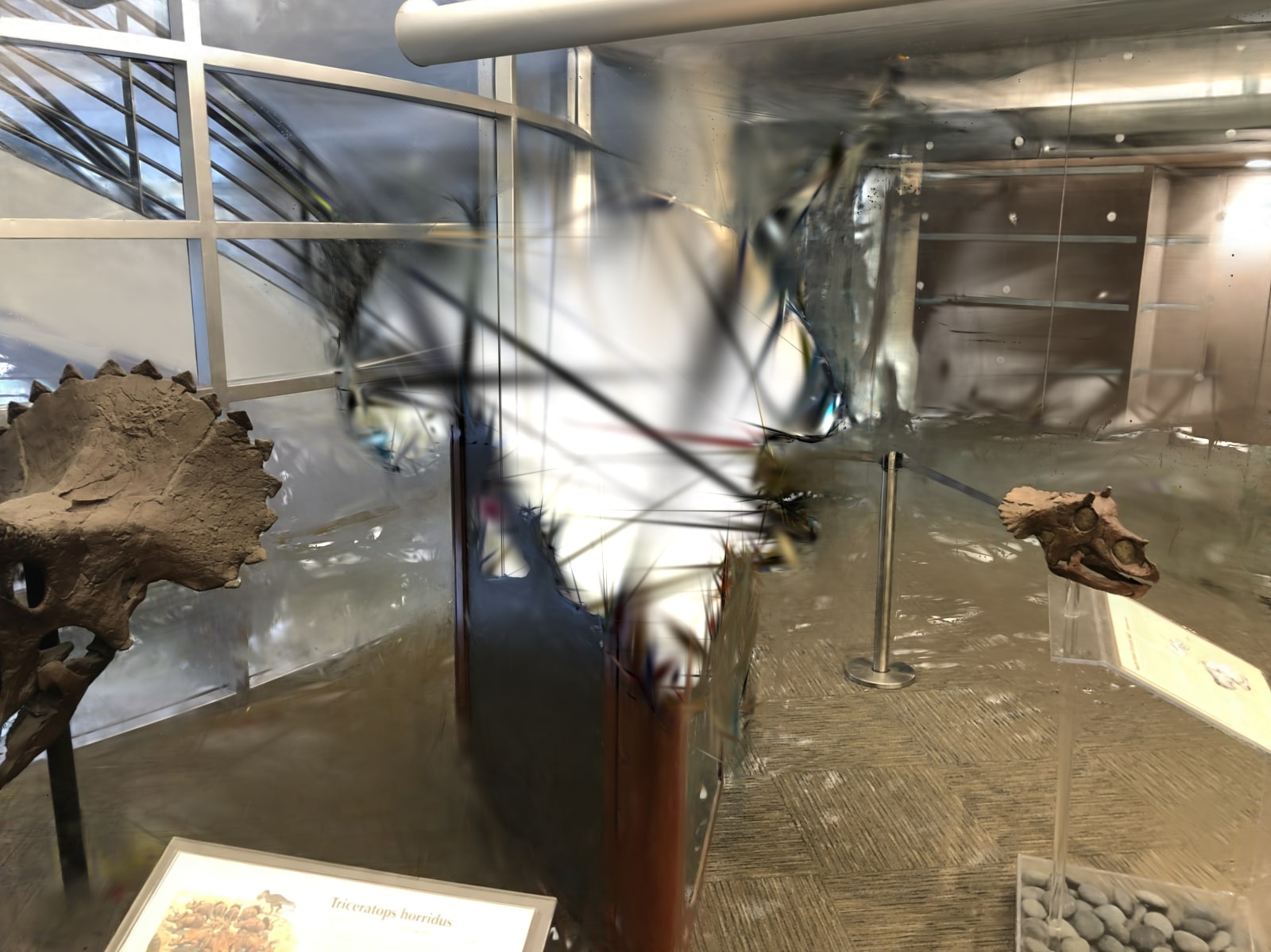}
\end{minipage}
\begin{minipage}{0.11\textwidth}
    \includegraphics[width=\linewidth]{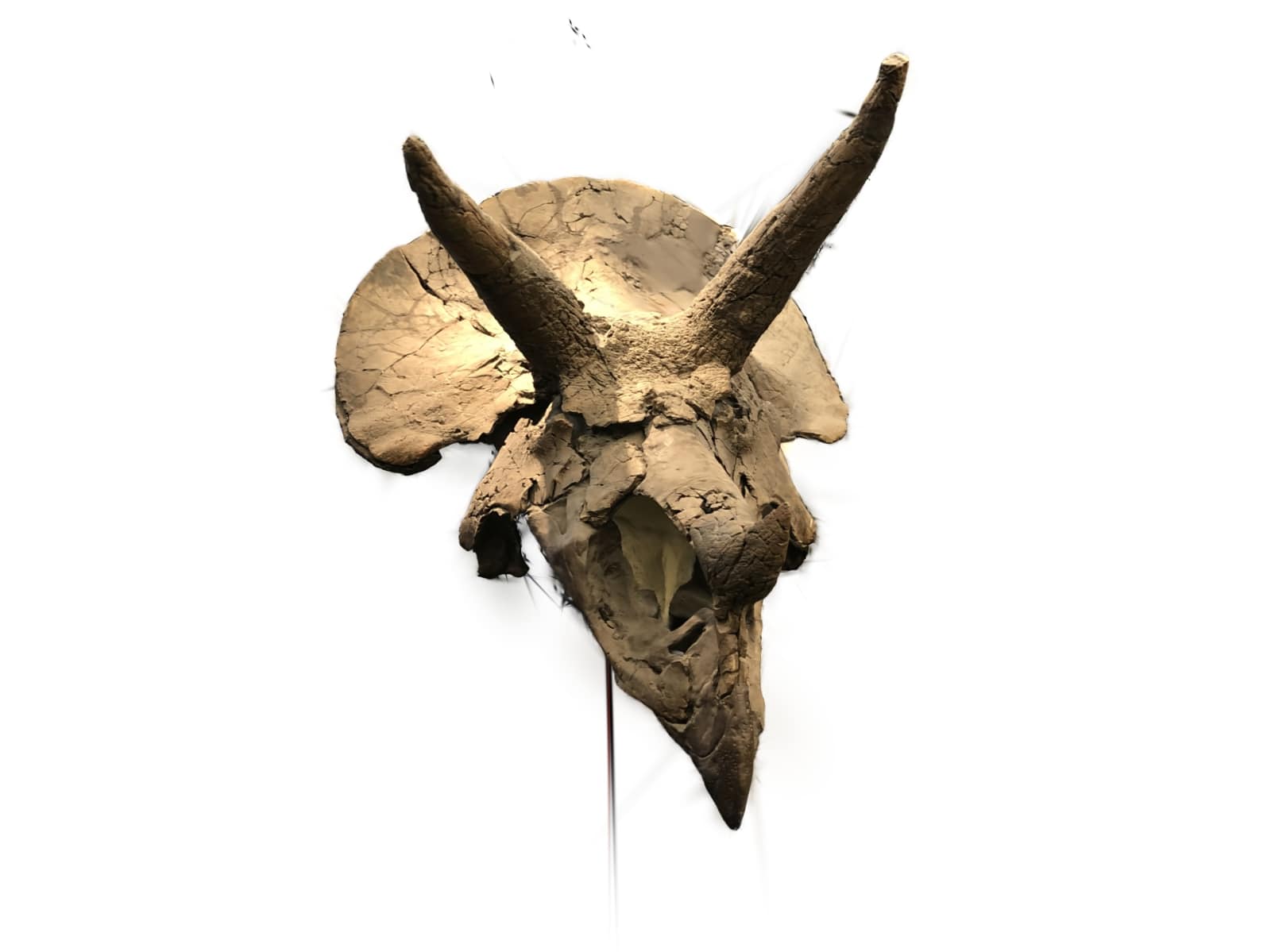}
\end{minipage} \\

\begin{minipage}{0.02\textwidth}
  \centering
  \rotatebox{90}{\textbf{teatime}}
\end{minipage}
\begin{minipage}{0.11\textwidth}
    \includegraphics[width=\linewidth]{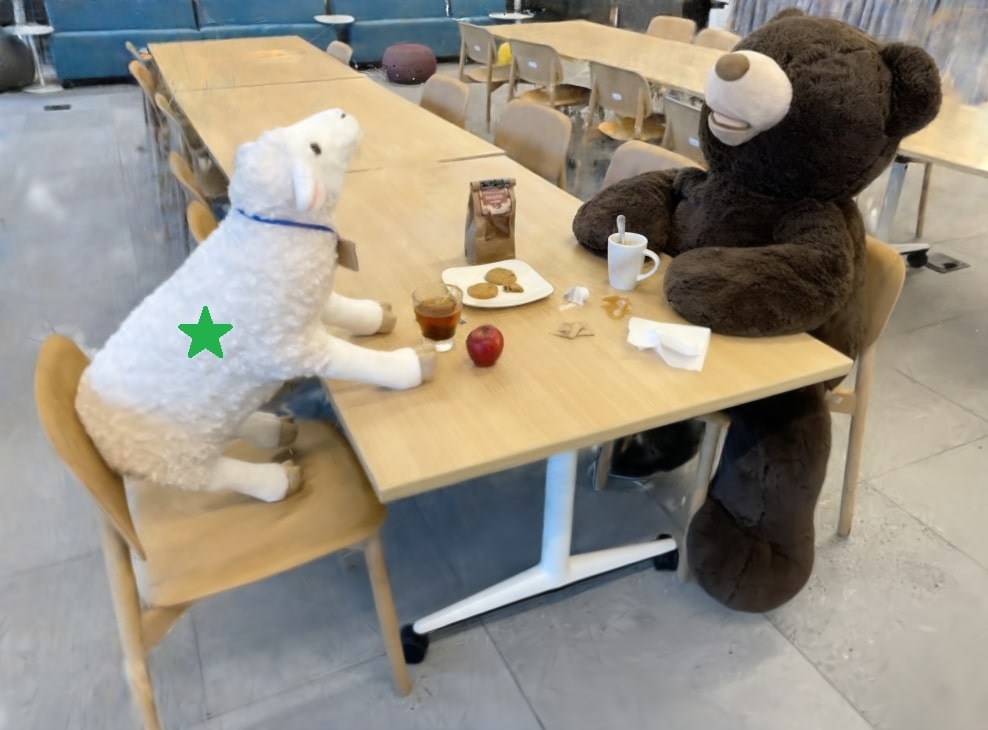}
\end{minipage}
\begin{minipage}{0.11\textwidth}
    \includegraphics[width=\linewidth]{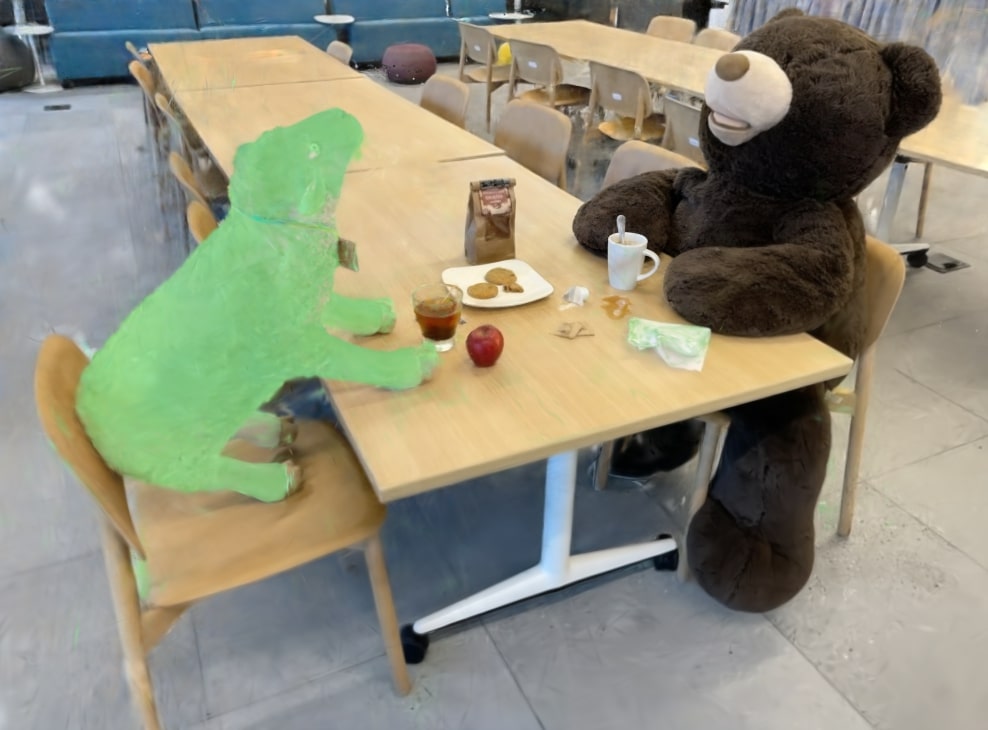}
\end{minipage}
\begin{minipage}{0.11\textwidth}
    \includegraphics[width=\linewidth]{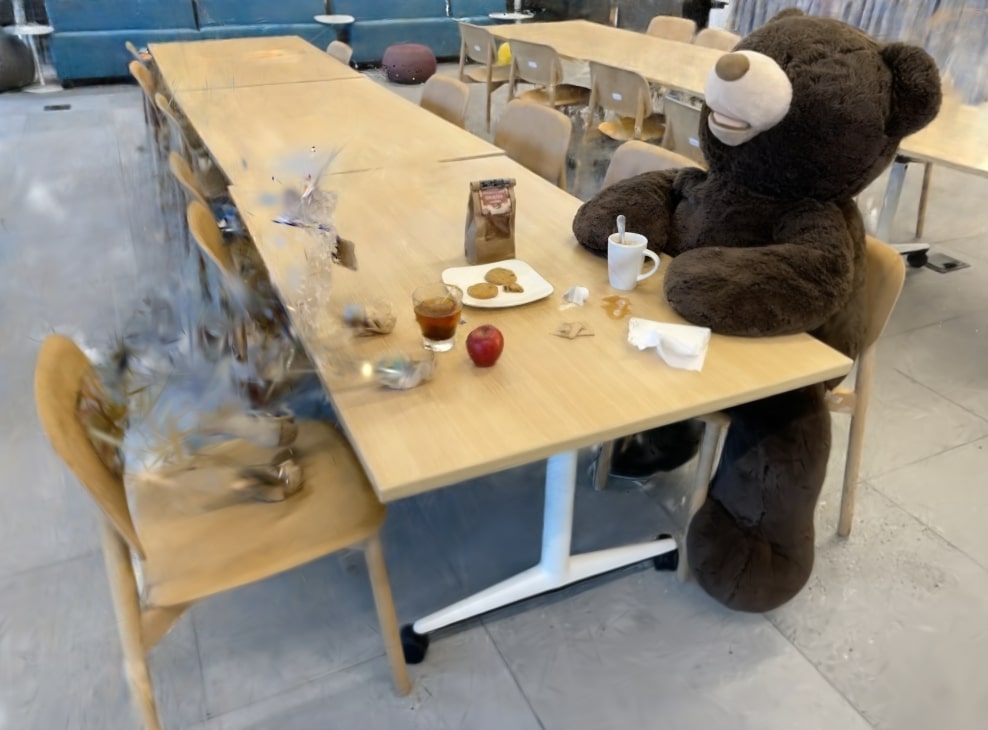}
\end{minipage}
\begin{minipage}{0.11\textwidth}
    \includegraphics[width=\linewidth]{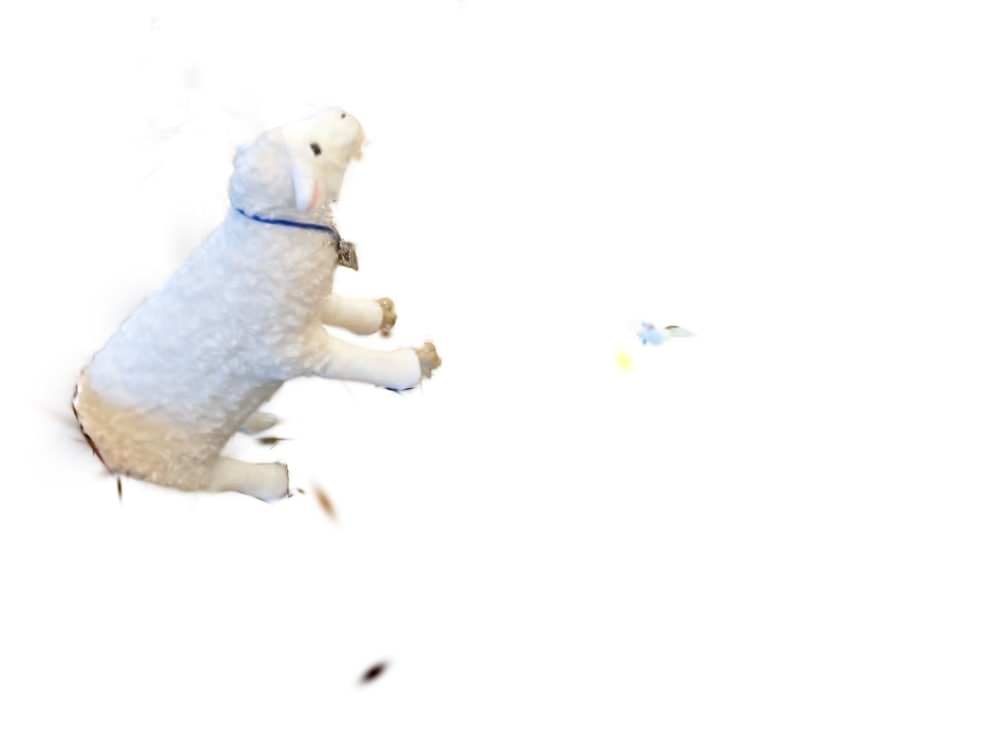}
\end{minipage}  \\

\begin{minipage}{0.02\textwidth}
  \centering
  \rotatebox{90}{\textbf{fortress}}
\end{minipage}
\begin{minipage}{0.11\textwidth}
    \includegraphics[width=\linewidth]{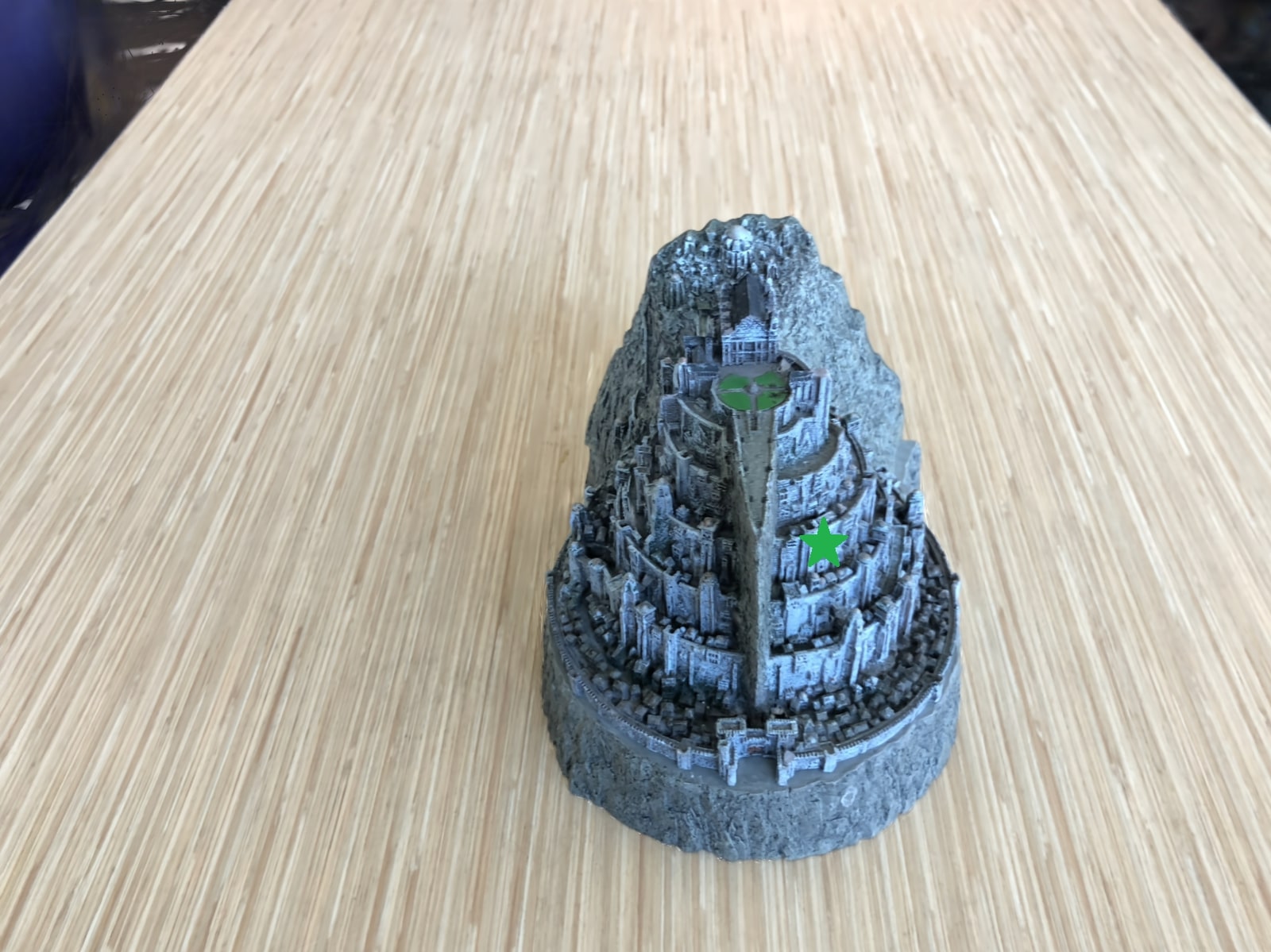}
\end{minipage}
\begin{minipage}{0.11\textwidth}
    \includegraphics[width=\linewidth]{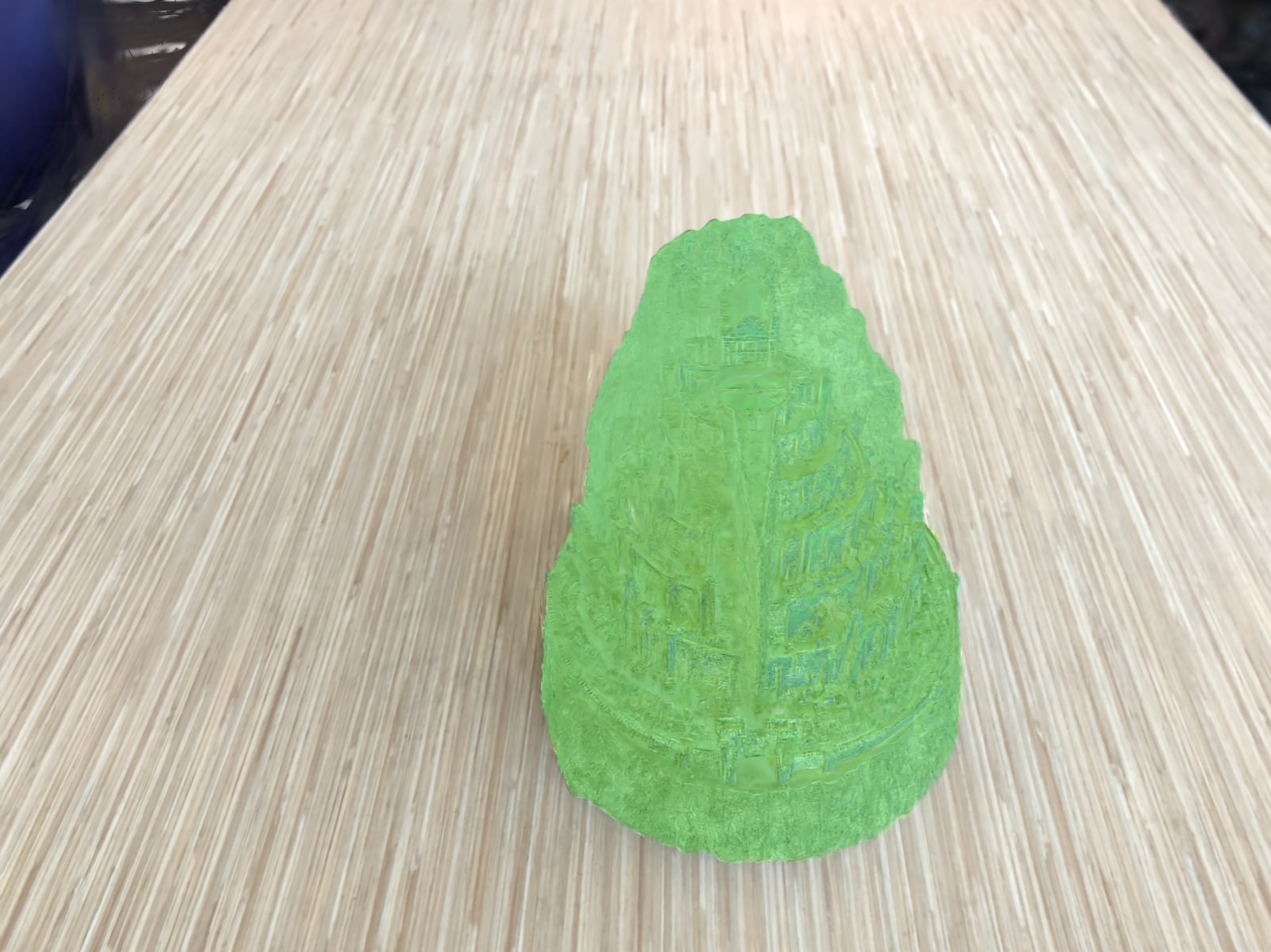}
\end{minipage}
\begin{minipage}{0.11\textwidth}
    \includegraphics[width=\linewidth]{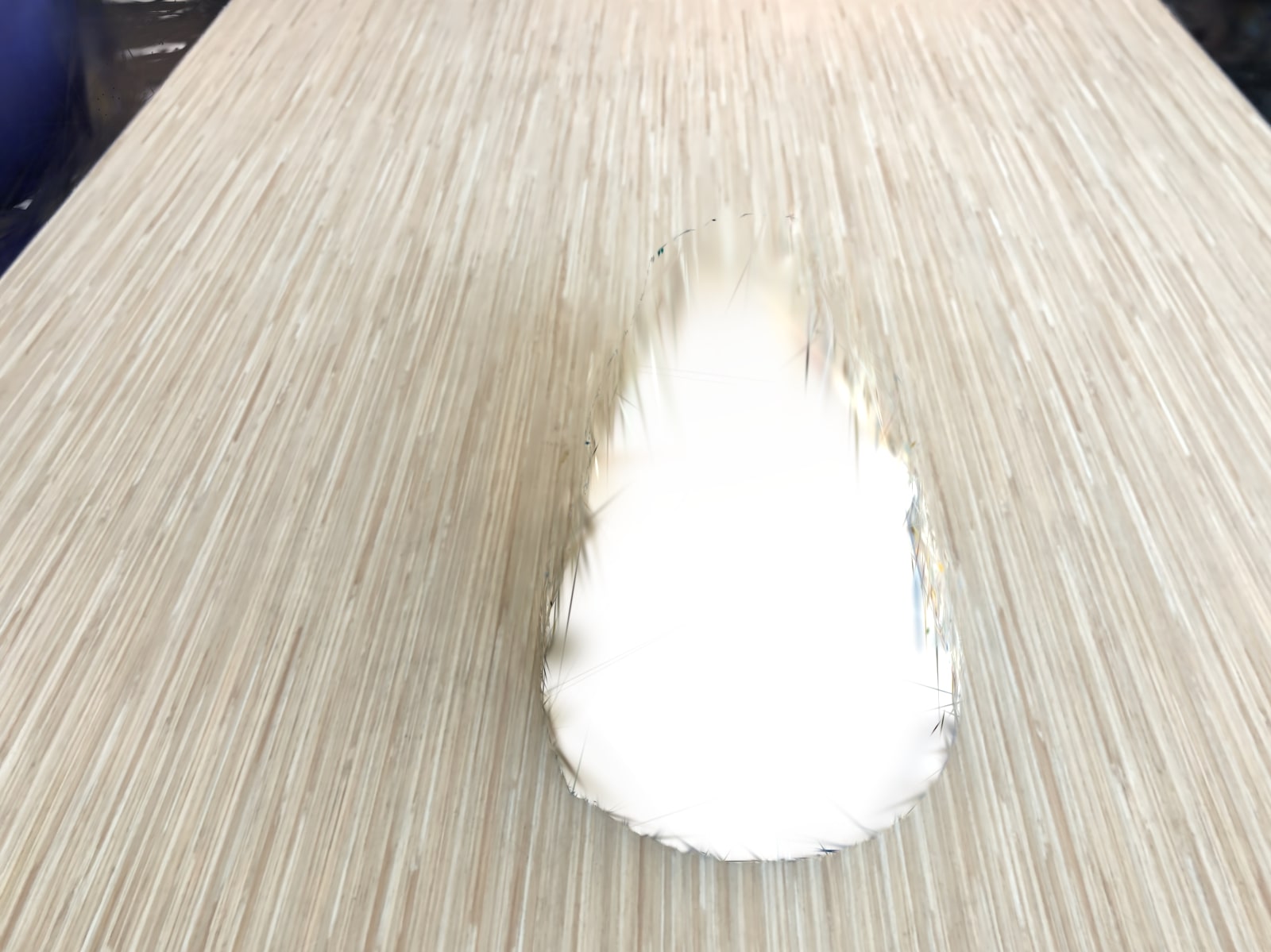}
\end{minipage}
\begin{minipage}{0.11\textwidth}
    \includegraphics[width=\linewidth]{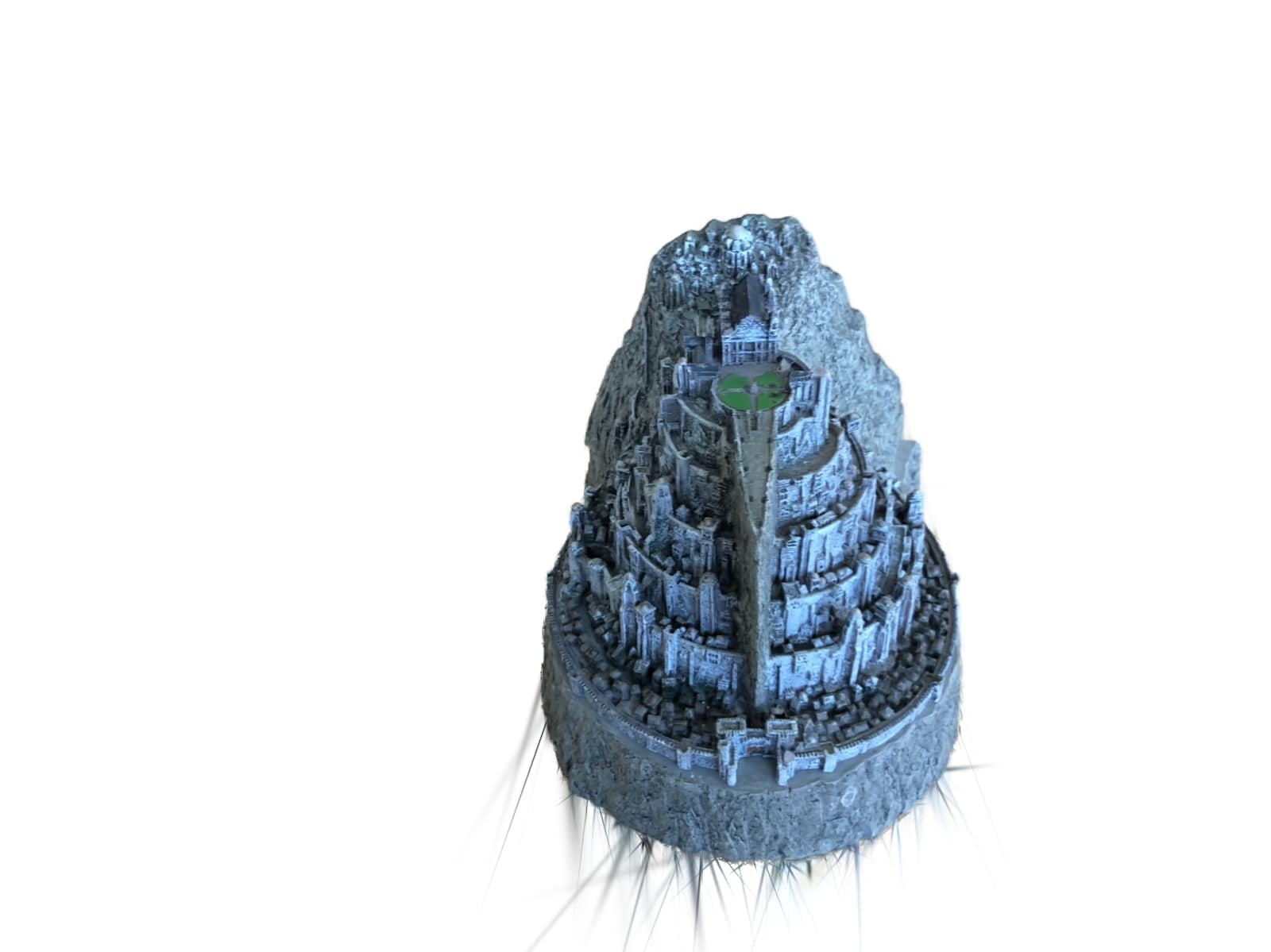}
\end{minipage} 

\caption{Qualitative results of 3D scene editing tasks on scenes, showing sharp object boundaries and high visual consistency.}
\label{fig:scene_editing_task}
\end{figure}
\begin{figure}[t!]
\renewcommand{\arraystretch}{0.9}
\small
\centering
\begin{minipage}{0.45\textwidth}
    \includegraphics[width=\linewidth]{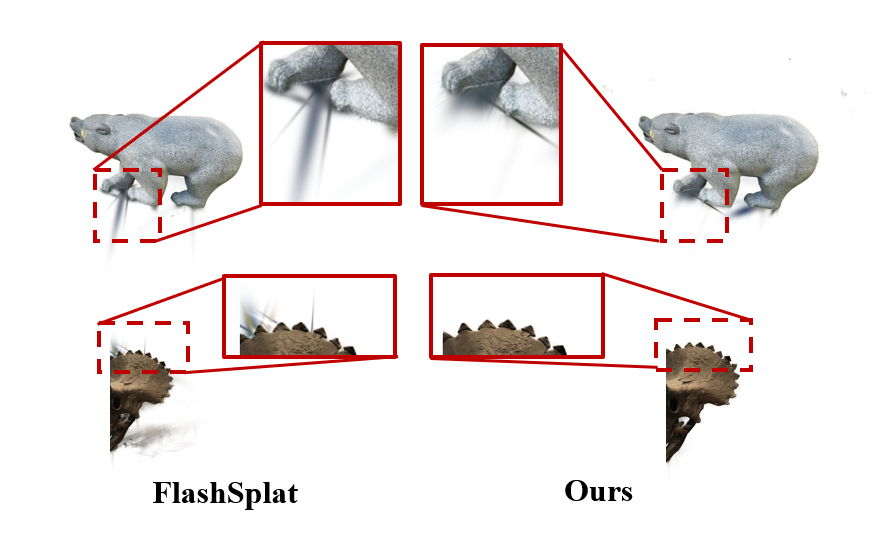}
\end{minipage}
\caption{Comparison of object removal quality where FlashSplat produces large artifacts and blurry boundaries, whereas our method achieves significantly cleaner and sharper object boundaries. [Top : bear scene, Bottom: horns scene]}
\label{fig:scene_quality_comparion_task}
\end{figure}
\vspace{-1mm}
\subsection{Quantitative Results}
The quantitative results in Table~\ref{tab:reconstruction_metrics_results} demonstrate the effectiveness of our preprocessing pipeline for noise removal. Thus, our high-quality masks yield superior scene fidelity compared to SAM~\cite{Kirillov_2023_ICCV} masks used by previous methods~\cite{ye2025gaussiangrouping, shen2025flashsplat}. Table~\ref{tab:method_result_comparison} shows a comparable segmentation accuracy (slightly lower) on the NVOS~\cite{Ren_2022_CVPR} dataset because the provided SAM~\cite{Kirillov_2023_ICCV} masks exhibit ambiguities and boundary inconsistencies, degrading the learned features. In contrast, results from the Table~\ref{tab:scene_wise_results_comparison} shows that our SAM-HQ + noise removal pipeline outperforms FlashSplat (mIoU 80.52\% vs 80.25\%), demonstrating the importance of mask quality for robust downstream editing tasks. Our pipeline achieves real-time rendering speed (average 115 FPS across scenes on a mid-range GPU) while enhancing segmentation accuracy.
\vspace{-2mm}
\subsection{Qualitative Results}
Fig.~\ref{fig:object_mask_extraction} illustrates the improved quality of object masks after our noise removal pipeline removes small artifacts from SAM-HQ~\cite{ke2023segment} generated masks. Additionally, Fig.~\ref{fig:scene_editing_task} demonstrates the effectiveness of our robust approach in interactive editing tasks, including object recoloring, removal, and extraction, guided by user-specified 3D point prompts. Fig.~\ref{fig:scene_quality_comparion_task} shows that our method produces noticeably cleaner and more precise boundaries. This improvement stems primarily from the use of high-quality (HQ) segmentation masks.

\begin{figure}[t!]
\renewcommand{\arraystretch}{0.9}
\small
\centering
\hspace{-0.1cm}\textbf{$\gamma_{\text{p}}$ = $-$1.0}  \hspace{1.4cm}\textbf{$\gamma_{\text{p}}$ = 0.2}  \hspace{1.5cm}\textbf{$\gamma_{\text{p}}$ = 1.0} \\

\begin{minipage}{0.15\textwidth}
    \includegraphics[width=\linewidth]{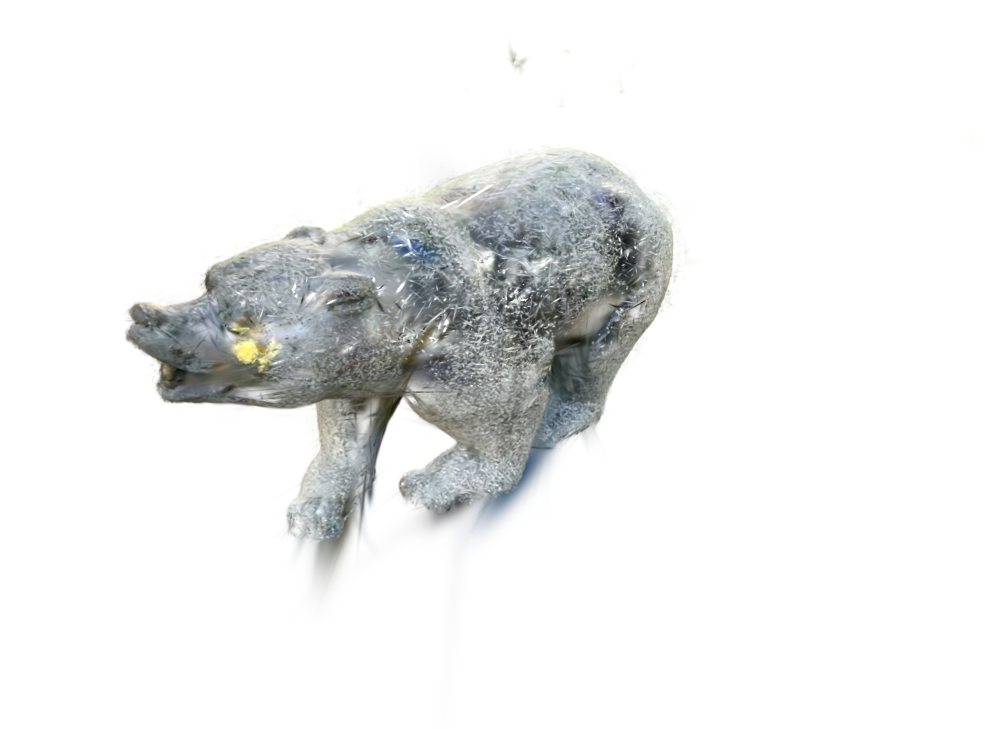}
\end{minipage} 
\begin{minipage}{0.15\textwidth}
    \includegraphics[width=\linewidth]{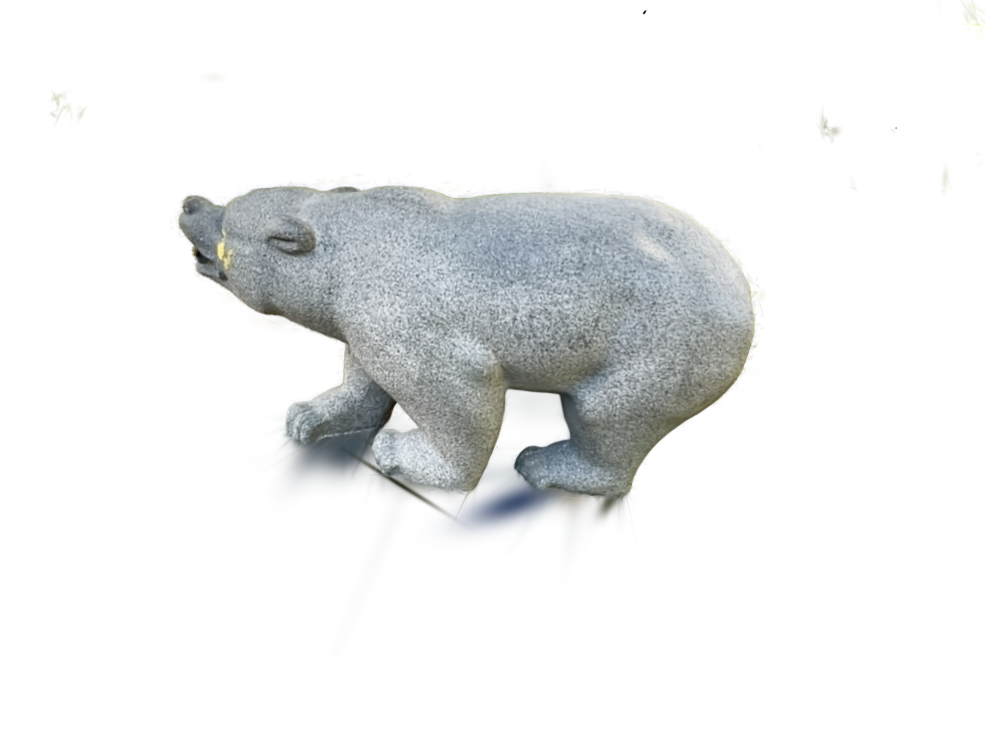}
\end{minipage} 
\begin{minipage}{0.15\textwidth}
    \includegraphics[width=\linewidth]{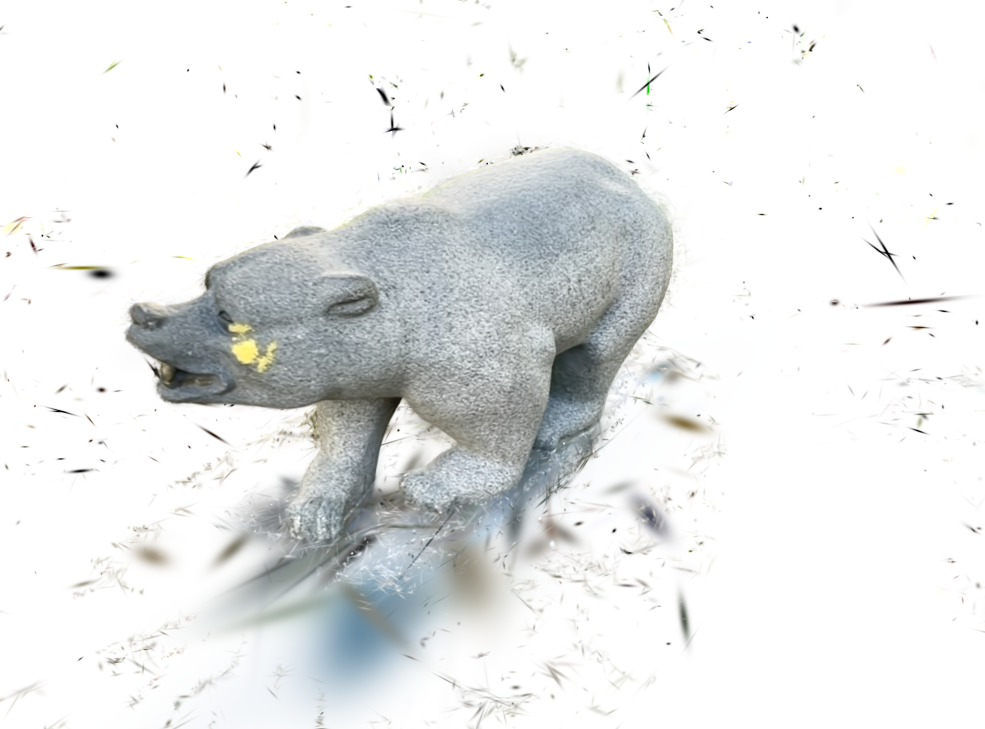}
\end{minipage} \\
\begin{minipage}{0.15\textwidth}
    \includegraphics[width=\linewidth]{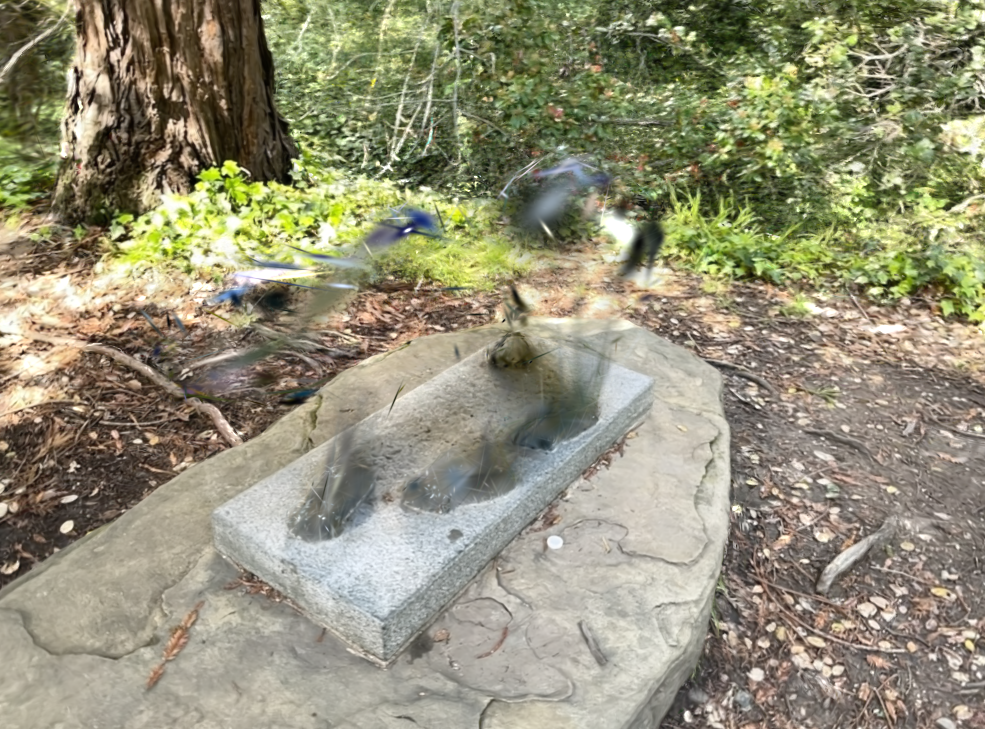}
\end{minipage} 
\begin{minipage}{0.15\textwidth}
    \includegraphics[width=\linewidth]{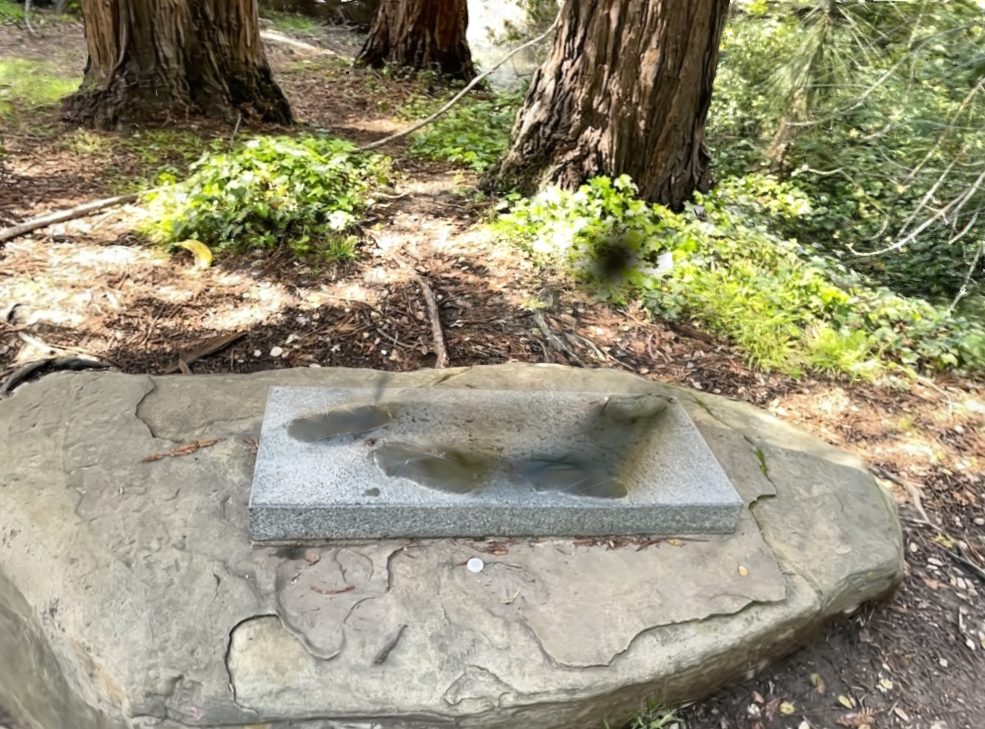}
\end{minipage}
\begin{minipage}{0.15\textwidth}
    \includegraphics[width=\linewidth]{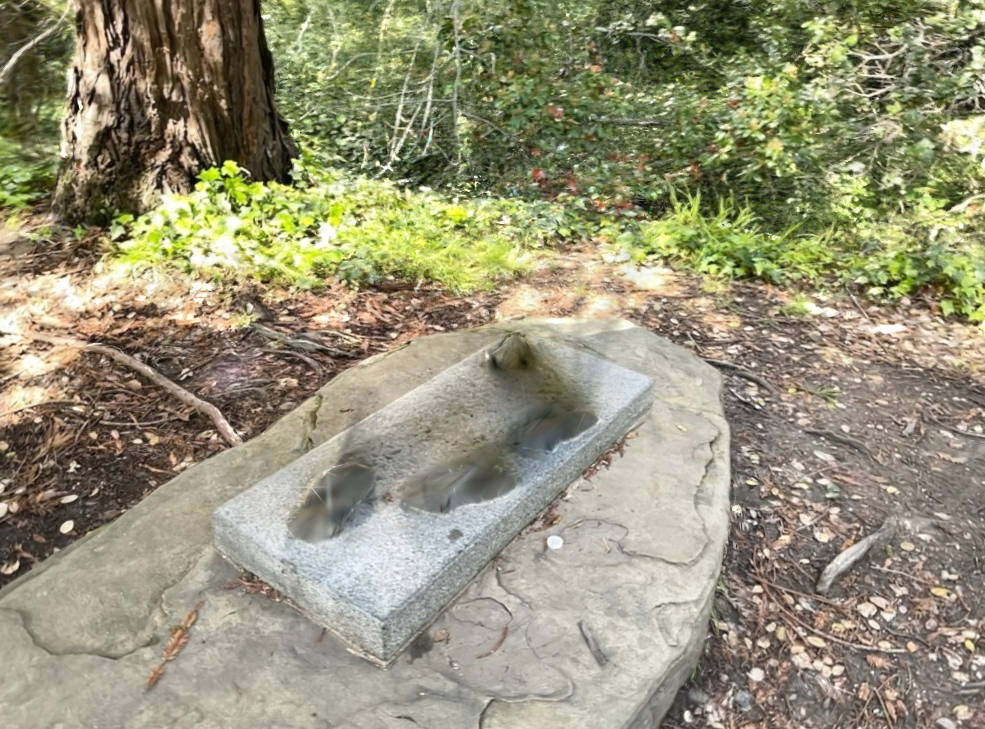}
\end{minipage}  \\

\caption{Effect of varying $\gamma_{\text{p}}$ which controls the contribution of prior learned labels during label reassignment for scene bear.}
\label{fig:ablation_varying_prior_value}
\end{figure}
\subsection{Ablation Studies}
To investigate the impact of the prior weight of the learned label $\gamma_{\text{p}}$ on the label reassignment framework, we conducted a qualitative ablation study on the scene bear, focusing on the object removal task as shown in Fig.~\ref{fig:ablation_varying_prior_value}. We varied $\gamma_{\text{p}}$ across values of $-$1.0, 0.2 (default) and 1.0, and evaluated the quality of both the remaining scene and the removed object. In the negative domain, $\gamma_{\text{p}}$ = $-$1.0 degrades performance, as some Gaussians persist in the remaining scene, introducing artifacts in both the remaining scene and the removed object. Conversely, in the positive domain, $\gamma_{\text{p}}$ = 0.2 (default) yields a smoother remaining scene and removed object by balancing confident priors, but increasing to $\gamma_{\text{p}}$ = 1.0 introduces multiple artifacts in the removed object due to incorrectly included outlier Gaussians. Our study demonstrates the importance
of carefully tuning the prior weight $\gamma_{\text{p}}$ to balance prior influence according to different use cases in our 3D Gaussian Splatting segmentation framework.

\section{Conclusion}
In this paper, we present a robust prior-guided 3D segmentation method for 3D Gaussian Splatting (3D-GS) scenes, leveraging learned object label priors derived from joint training with object features. To achieve precise boundary segmentation, we developed a preprocessing pipeline with a noise removal module to generate high-quality, view-consistent masks. Experiments on scenes from the LeRF, Mip-NeRF, and LLFF datasets demonstrate the effectiveness of our approach. Beyond SAM-HQ integration, our key innovation lies in eliminating heuristic biases through prior-guided reassignment and joint optimization of view-invariant priors, enabling multi-view consistency and interactive editing with reduced artifacts. Although our work focuses on robust segmentation for object removal in 3D Gaussian Splatting (3D-GS), we intentionally left the voids unfilled to showcase the raw segmentation strength. Our work can be expanded in the future by adding generative inpainting, such as diffusion-based models, to fill the void spaces naturally after object editing. This would create more realistic and seamless edited scenes, making the framework even better for real-world interactive applications like 3D scene editing or AR/VR experiences.


\bibliographystyle{IEEEbib}
\bibliography{strings,refs}
\end{document}